\documentclass[preprint,12pt]{elsarticle}

\usepackage{graphicx}            
\usepackage{amsmath,amssymb}     
\usepackage{siunitx}             
\usepackage[caption=false]{subfig} 
\usepackage[dvipsnames]{xcolor}

\usepackage[linesnumbered,ruled,vlined]{algorithm2e} 
\usepackage{amsthm}
\usepackage{adjustbox}
\usepackage[normalem]{ulem}
\usepackage{url}
\usepackage[colorlinks = true,
            linkcolor = blue,
            urlcolor  = blue,
            citecolor = blue,
            anchorcolor = blue]{hyperref}
\SetKwComment{Comment}{$\triangleright$\ }{}

\SetCommentSty{mycommfont}
\SetKwInput{KwTrigger}{Trigger}
\newcommand{\Transmit}[1]{\textbf{Transmit:} #1}

\usepackage{amsthm}
\theoremstyle{plain}
\newtheorem{theorem}{Theorem}[section]

\theoremstyle{definition}
\newtheorem{definition}[theorem]{Definition}

\theoremstyle{remark}

\DeclareMathOperator{\sech}{sech}
\DeclareMathOperator{\sign}{sign}

\journal{Robotics and Autonomous Systems}

\begin{document}

\begin{frontmatter}

\title{Consensus Building in Human-robot Co-learning via Bias Controlled Nonlinear Opinion Dynamics and Non-verbal Communication through Robotic Eyes\tnoteref{t1}}

\tnotetext[t1]{This research was partially supported by NSF Grant ECCS-2218517 and was approved by the Institutional Review Board (IRB \#2125317-1) at George Mason University, Fairfax, Virginia, USA.}

\author[gmuece]{Rajul Kumar}
\author[gmuece,potomac]{Adam Bhatti}
\author[gmuece]{Ningshi Yao\corref{cor1}}

\cortext[cor1]{Corresponding author}
\ead[cor1]{nyao4@gmu.edu}

\affiliation[gmuece]{organization={Department of Electrical and Computer Engineering, George Mason University}, 
            addressline={4400 University Drive}, 
            city={Fairfax},
            postcode={22030}, 
            state={Virginia},
            country={USA}}

\affiliation[potomac]{organization={The Potomac School}, 
            city={McLean},
            state={Virginia},
            country={USA}}

\begin{abstract}
Consensus between humans and robots is crucial as robotic agents become more prevalent and deeply integrated into our daily lives. This integration presents both unprecedented opportunities and notable challenges for effective collaboration. However, the active guidance of human actions and their integration in co-learning processes, where humans and robots mutually learn from each other, remains under-explored. This article demonstrates how consensus between human and robot opinions can be established by modeling decision-making processes as non-linear opinion dynamics. We utilize dynamic bias as a control parameter to steer the robot’s opinion toward consensus and employ visual cues via a robotic eye gaze to guide human decisions. These non-verbal cues communicate the robot’s future intentions, gradually guiding human decisions to align with them. To design robot behavior for consensus, we integrate a human opinion observation algorithm with the robot’s opinion formation, controlling its actions based on that formed opinion. Experiments with $51$ participants ($N=51$) in a two-choice decision-making task show that effective consensus and trust can be established in a human–robot co-learning setting by guiding human decisions through nonverbal robotic cues and using bias-controlled opinion dynamics to shape robot behavior. Finally, we provide detailed information on the perceived cognitive load and the behavior of robotic eyes based on user feedback and post-experiment interviews.
\end{abstract}

\begin{highlights}
\item Pioneered the application of dynamic bias in non-linear opinion dynamics to facilitate consensus in human–robot interaction, utilizing robotic eye gaze as a non-verbal communication channel to convey the robot’s intent to human counterpart.

\item Designed a bi-directional decision-making framework that transitions from initial robot–human dissensus to consensus through visual cues, validated experimentally with 51 human participants.

\item Demonstrated that humans' trust in robotic guidance progressively increases with repeated exposure to consistent visual cues, evidenced by reduced hesitation behaviors and higher direct consensus rates.
\end{highlights}


\end{frontmatter}


\section{Introduction}\label{Sec_1}
For the modern industry, achieving safe and efficient human-robot collaboration is essential to maximize productivity, improve workplace safety, and realize the full potential of robotic technologies. In collaborative scenarios, humans and robots naturally engage in a co-learning process, mutually adapting their behaviors based on each other's actions \cite{Zoelen2021HRI, Young2019CDC}. However, the dynamics of opinion formation in such human-robot co-learning process remain insufficiently understood, making it challenging to design robot behaviors that reliably reach consensus with human actions. As a result, robots often remain confined within safety cages, reinforcing a physical separation between robotic and human workspaces. However, even in routine industrial assembly tasks, critical thinking is often required, highlighting the need for human oversight and collaboration since robots alone are insufficient \cite{Pfeiffer2016}. While robots excel at repetitive tasks, humans provide essential decision-making skills in uncertain environments. 
Thus, integrating human cognitive abilities with robotic precision through human-in-the-loop systems is essential, viewing these capabilities as complementary rather than competitive. Effective integration hinges on achieving consensus, defined as the state where human and robot agents align on shared goals or opinions. Without consensus, collaboration can become inefficient or hazardous due to conflicting objectives or misinterpreted intentions. Therefore, advancing human-robot collaboration requires a deeper exploration into the dynamics of opinion formation and the development of robust consensus-building mechanisms within human-in-the-loop frameworks.

Existing literature frequently neglects the critical role of human input in consensus-building processes by treating human actions merely as uncertainties or noise in robot control algorithms \cite{Jaber2024ICPS}. In contrast, recent studies \cite{Cortigiani2025RAL, Losey2022, Williams2023}have begun to recognize human actions as direct inputs or states for robot control, typically with humans leading and robots responding. However, these approaches lack strategic guidance from robots and explicit communication of robot opinions to humans.  The consensus is fundamentally a two-way process that requires a mutually satisfactory reconciliation of human and robotic perspectives. Moreover, human behavior is inherently unpredictable and often irrational \cite{Johnson2021}. Without strategic guidance or communication, humans can engage in behaviors that trick or attempt to surpass the capabilities of their robotic counterparts. 

Achieving effective consensus in opinions and actions requires robot behavior to be designed with a clear understanding of closed-loop human–robot co-learning dynamics. Few studies have explicitly explored this aspect of co-learning \cite{Zoelen2021HRI, Kumar2024HRI, Zoelen2025RAL, Kumar2024ROMAN}. In particular, a recent work \cite{Zoelen2025RAL} proposed a Q-learning–based approach to facilitate co-learning in a human–robot handover task. Although this work represents a valuable step forward, the use of reinforcement learning limits the insight into mutual opinion formation dynamics between human and robot. In \cite{Young2019CDC}, the authors provided theoretical guarantees of consensus under specific parameter regimes, assuming the human mind follows the classical Rescorla-Wagner (or RW) decision-making model \cite{Rescorla1972}. However, our previous work \cite{Kumar2024HRI} presented real-world co-learning experiments in a two-choice decision-making task, revealing that human behavior can be highly unpredictable and shaped by cognitive biases, social attention, and diverse individual traits. Such complexities cannot be captured by an oversimplified RW model. To address this gap, we proposed a proactive robot-control approach that leverages transfer learning to build a data-driven mental model of human choices in \cite{Kumar2024ROMAN}, which enables the robot to anticipate the user’s future actions and facilitate consensus. Although the data-driven, machine-learning model proposed in \cite{Kumar2024ROMAN} can capture sudden irrationalities and can model diverse human decisions, it relies heavily on extensive human-robot interaction datasets and offers limited interpretability.

Addressing these limitations, we introduce a bias-controlled nonlinear opinion dynamics model based on recent frameworks \cite{Bizyaeva2023, Leonard2024}. Our model utilizes bias as a robot control parameter and utilizes robotic visual cues to guide human choices proactively. Unlike static models that focus only on the final decision, the opinion dynamics framework captures the temporal evolution of preferences, enabling immediate recognition and intervention in disagreements or agreements. In addition, it incorporates key psychological factors such as social attention, memory, and a bias term. The model’s analytical tractability also enables examination of its stability, convergence, and uniqueness properties of agent opinions. Building on this foundation, we introduce a dynamic bias control strategy, where the robot modulates its internal bias parameter to guide its opinion toward consensus with the human, making the human an integral part of the process. To ensure the human remains informed and engaged in the process, this bias is communicated non-verbally through robotic eye gaze, serving as an intuitive and socially meaningful cue. This human-in-the-loop design supports transparent intention sharing, improves collaboration fluency, and reduces the dependency on large task-specific datasets typically required by black-box machine learning models.

A large body of work exists on opinion-dynamics models in multi-agent interactions, such as the DeGroot model, the Friedkin--Johnson model, and the Hegselmann--Krause model \cite{Hegselmann2022JASSS}. However, many of these are linear in nature, and evidence suggests that human opinion formation is highly nonlinear \cite{Kim2008DSS, Andrzej2012JTACS}. Moreover, previous models often omit key psychological factors such as cognitive bias and social attention. Our selected nonlinear model \cite{Bizyaeva2023, Leonard2024} addresses these limitations, demonstrating superior mathematical tractability, flexibility, and robustness, making it ideal for human-robot two-choice consensus-building experiments.

In this article, we present a two-choice decision-making experiment involving a human and a robotic arm that can operate either cooperatively or competitively, to investigate how consensus emerges when both are modeled by our proposed bias-controlled nonlinear opinion dynamics. We first iteratively refined the model parameters by trial and error in the experimental settings described in section \ref{Sec_3}. Using these optimized parameters, we performed a numerical parametric sweep and equilibrium point analysis to identify the critical bias thresholds that allow a transition from dissensus to consensus. We then introduced a dynamic bias updating rule that continuously changes the robot’s bias toward one of the two choices. Finally, we demonstrate how real-time human opinions can be observed and guided through visual cues from a robotic eye, serving as an effective way to bias the human's opinion during interactions. Our contributions include:\\
1. We present a novel experimental study demonstrating how nonverbal cues from robotic eyes can deliberately bias human opinion, marking a significant step toward enhancing meaningful human participation in consensus-building.\\
2.	We introduce the first application of dynamic bias in non-linear opinion dynamics for consensus building, using anthropomorphic robotic eye gaze as a biasing mechanism. Furthermore, we investigate the correlation between the increase in visual cues from the robotic eye and the corresponding increase in human trust over time.\\
3.	For the presented human-robot interaction scenario, we tuned the parameter set of the non-linear opinion dynamics. Through parametric analysis, we characterized the system’s behavior under varying bias conditions and revealed the mechanisms behind the robot’s path to consensus. Notably, once the robot exceeds the maximum social influence of collective human opinions, dynamic self-biasing becomes the dominant force driving it toward consensus.\\
4.	Our pioneering experiment establishes a strategic bi-directional decision-making process between humans and robots, characterized by a tightly integrated co-learning closed loop. We investigate how repeated robot-driven disagreements affect human decision-making and whether these initial conflicts subsequently prompt human alignment with the robot’s visual cues.

The rest of this article is organized as follows. Section \ref{Sec_2} provides a comprehensive review of both past and recent related work. Section \ref{Sec_3} details the experimental setup, including components and the procedural methodology. Section \ref{Sec_4} presents the numerical analysis for biases in non-linear opinion dynamics, along with the proposed opinion formation and robot behavior algorithms. Section \ref{Sec_5} presents the incorporation of human in co-learning interactions with the robot through the use of robotic eye gaze. The subsequent Sections~\ref{Sec_6}, \ref{Sec_7}, and \ref{Sec_8} present experimental demonstrations, overall results, and insights from user feedback, respectively.


\section{Related Work}\label{Sec_2}
This section reviews the relevant literature on robot control design, nonverbal communication, human behavior regulation, and experimental design for colearning in human-robot interactions.
\subsection{Robot Control Design for Human-Robot Consensus Building}\label{Sec_2.1}
Effective human-robot collaboration is significantly dependent on trust, fostered through safe and proactive robot decision-making capabilities. Previous research frequently employs Markov Decision Processes (or MDPs) \cite{Unhelkar2020, Stewart2012} and Partially Observable Markov Decision Processes (or POMDPs) \cite{Losey2022, Williams2023} for robot control.  Although these methods robustly handle uncertainty, they typically neglect individual human differences and rely on abstract environmental representations, making robot actions less intuitive. Consequently, robot actions may seem unpredictable to human collaborators. In contrast, opinion dynamics explicitly incorporates observable social and psychological factors into robot actions and models other agents' actions as opinions, considering individual variability, thereby enhancing predictability and trust. This improves the intuitiveness and adaptability of robotic behavior, essential for consensus building.

The pioneering application of non-linear opinion dynamics for robot navigation control, as presented in \cite{Cathcart2023}, addresses collision avoidance in confrontational human-robot navigation scenarios. Although innovative, the approach is unilateral, assigning all collaborative navigation responsibilities to the robot and positioning it as the sole active participant, thus limiting human involvement. Furthermore, in situations of persistent disagreement between agents, Section~V.C of \cite{Bizyaeva2023} explores the transition from disagreement to agreement by dynamically modulating the inter-agent coupling weight on a shared opinion. 
However, directly communicating the inter-agent coupling weight—defined as the robot’s influence on human decisions—as a control signal in real-world scenarios is both challenging and impractical. In contrast, biases or external stimuli, which are comparatively easier to quantify, can be effectively conveyed to humans through visual cues such as eye gaze directed toward a specific option, offering a feasible alternative~\cite{Onnasch2023}.

To the best of the authors' knowledge, this is the first study to incorporate bias into robot actions, and external stimuli from robot to human, exploring collaborative outcomes by influencing human opinions with bias, thereby making the collaborative process bidirectional in building consensus.

\subsection{Non-verbal Communication: Robotic Eye Gaze}\label{Sec_2.2}
Human opinion formation is influenced by various psychological factors that are not yet fully understood; integrating these factors, such as continuously evolving biases and opinions, into robot behavior presents a significant challenge. Designing robot autonomy in interactive scenarios is inherently challenging, especially when robots are expected to simultaneously guide human decisions through non-verbal cues and exhibit human-like decision-making abilities. Non-verbal cues, such as eye gaze, are rapid, intuitive, and effective for conveying intent~\cite{kalpagam2018RAM, mendez2023H}. Unlike verbal communication, which can be slow, prone to misinterpretation, or perceived as rude depending on the robot's tone, nonverbal cues such as eye gaze play a crucial role in reducing confusion in real-time interactions, making them swift and intuitive~\cite {Mavridis2015, Admoni2017, Zhao2015,  mendez2023H}. Their clarity makes them invaluable for conveying intent and building rapport in human-robot co-learning. Moreover, to meet human expectations in collaborative settings, robots require human-like decision-making capabilities that make their behavior more transparent, predictable, and socially acceptable to human partners \cite{Dragan2015HRI, Breazeal2003IJHCS, Waytz2014JESP}. A robust theoretical understanding of how robotic cues influence human opinions is required, particularly in their role as external stimuli or biases that can drive a team toward either consensus or dissensus. 

Robotic eye gaze, in particular, serves as a predictive and influential cue. Research works in~\cite{Zhao2015, Kiilavuori2021} demonstrate that humans instinctively adopt the visual perspective of a robot. In addition, studies on self-driving cars have shown that anthropomorphic robotic eyes significantly reduce risky pedestrian behaviors by clearly communicating vehicle turning intentions~\cite{Chang2022AutomotiveUI, Gui2022AutomotiveUI, Wang2023MyEyes}.  In human-to-robot handover scenarios, well-designed gaze patterns enhance likability and communicative effectiveness~\cite{Faibish2022IJSoR}, with subtle differences in gaze patterns that dramatically affect perceived naturalness~\cite{Yuguchi2021Android}.

Additionally, studies indicate that monitoring human eye gaze can accurately predict user intentions and actions, thus enhancing collaboration \cite{Fathaliyan2018Frontiers, Wang2020Frontiers, Boucher2012Frontiers}.  However, many existing studies rely on screen-based or virtual manipulations \cite{Onnasch2023, Psarakis2025}, which may not translate well into real-world interactions \cite{Shinozawa2005}. Consequently, recent research emphasizes the development of physical robotic eyes capable of lifelike movements and expressive states \cite{Pencic2022MDPI, Pencic2022Sensors, Breazeal2000}.

Building upon these insights, our study leverages the 3D robotic eye gaze in a decision-making context to investigate how varying gaze cues impact trust and consensus between humans and robots. Specifically, we explore how dynamically evolving robotic visual communication intentionally shapes human opinions toward consensus.

\subsection{Human Behaviour Regulation}\label{Sec_2.3}
Recent studies primarily utilize haptic feedback \cite{Grushko2021} or combined haptic-visual cues \cite{Habibian2023} to regulate human behavior. Although effective in communicating movement intentions, these approaches face scalability issues in complex industrial environments. Alternative expressive approaches, combining facial, body, and vocal signals \cite{Breazeal2002}, foster synchronization and alignment, but sustaining nuanced interactions over extended periods remains challenging, particularly with robots lacking anthropomorphic expressiveness.

Research further indicates that overly dominant robotic behaviors can be counterproductive \cite{Shane2021SR}. Hence, our approach positions the robot as a collaborative peer, employing precise nonverbal eye-gaze cues to influence human decisions equitably, fostering genuine consensus-building and avoiding the pitfalls associated with authoritative robotic interactions.

\subsection{Co-learning Experimental Design}\label{Sec_2.4}
Numerous experimental frameworks have been designed to validate robotic algorithms and human mental models within simulated real-world environments. The study in \cite{vanDijk2023} details an experiment in which humans and robots are tasked with placing tiles in color-matched slots. However, this study, along with \cite{Howell2023}, restricts the variability of decisions during a trial, failing to capture the nuanced complexity of real-world interactions. Furthermore, \cite{Kwon2020} introduces a Collaborative Cup Stacking Task where neither humans shift their strategies from less to more stable cup stacking configurations during a trial, nor do robots guide them to enhance performance, impacting the task's optimal outcome. Despite extensive research on intelligent decision-making models, many experimental frameworks delegate only strategic decision-making tasks to humans, relegating robots to logistical roles such as retrieving and supplying assembly parts \cite{Darvish2021}. This does not fully leverage robot capabilities, leading to under-utilization of their potential in collaborative tasks. In contrast, the study by \cite{Li2024} demonstrates humans performing agile tasks using VR technology while robots manage hazardous operations coordinated via a mutual cognitive system. However, the broad application of these methods is limited by the impracticality of using specialized virtual reality (or VR) devices in everyday industrial settings.

In contrast, our experimental design emphasizes dynamic, real-time adaptability and strategic interaction. Humans can modify their preferences during trials, and robots adapt proactively, focusing on strategic decision-making rather than logistic execution. Our framework integrates external psychological factors such as cognitive load, bias, and performance motivation, simulating complex social behaviors through initial disagreements that evolve toward consensus, differentiating our approach from static, controlled interaction experiments.

\section{Method : Experimental Setup and Procedure}\label{Sec_3}
We designed a controlled yet realistic experimental framework in which a human participant and a robot each select from two options, requiring mutual cooperation across repeated interactions to achieve steady-state consensus while simultaneously meeting individual game objectives. The following sections introduce the experimental setup, procedures, participant selection, and methodological considerations.
\subsection{Experimental Setup}\label{Sec_3.1}
\begin{figure}[t]
\centering   
\vspace{-0.5em}\includegraphics[width=\linewidth]{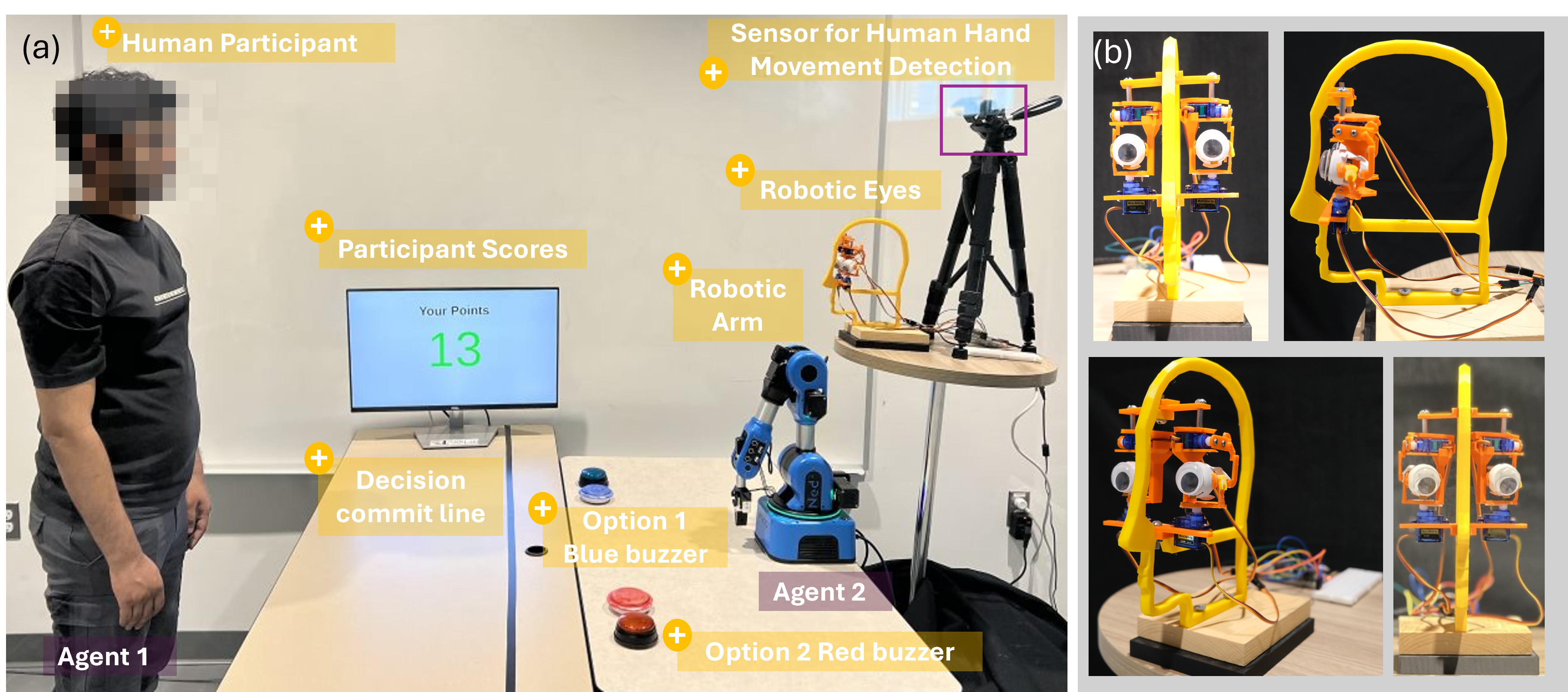}\vspace{-0.5em}
\caption{ (a) Experimental setup for a two-agent, two-choice decision-making task between a human and a robot (with human consent obtained for image use). The setup includes red and blue buzzers as two options for the human and robot to press, a robotic eye, a camera sensor, and a screen that continuously displays the participant's score.  (b) Robotic eye apparatus for nonverbal communication via eye gaze.}\vspace{-0.5em}
\label{fig:1}
\end{figure}

Fig.~\ref{fig:1}(a) illustrates the experimental setup, featuring a human participant and a robotic arm as two agents interacting with each other, where each agent has two options, i.e., press between a red or a blue buzzer. Participant's performance is evaluated in each trial, with a nearby screen displaying their real-time scores. Once participants cross the black line—referred to as the “decision commit line”—they are instructed not to change their choice, ensuring steady-state decision-making and maintaining safety by preventing last-minute adjustments.

To non-verbally indicate its intended choice through human-like eye gaze, the robotic eyes shown in Fig.~\ref{fig:1}(b) were mounted behind the arm. The robotic eye consists of two 3D-printed eyeballs, each driven by micro-servos allowing $180^\circ$ of rotation, allowing each eyeball to rotate $90^\circ$ to the left, right, up, or down from a neutral center position. Both eyeballs can be maneuvered independently or in unison to direct their gaze in any desired direction. To maintain a visually realistic appearance and enhance the humanoid features of the assembly, a 3D-printed nose and mouth structure was also integrated between the motorized eyeballs. More details on the robotic eye functioning can be found in our previous work \cite{huang2023see}. An overhead camera tracked the participant’s hand during buzzer selection, capturing dynamic transient and steady-state movements. And a synchronized speaker, programmed to emit the auditory cue ``1, 2, 3, go", was positioned behind the robotic eye and connected to a central control station, allowing synchronized execution of the robotic arm, eyes, and camera sensors. 
\subsection{Procedure}\label{Sec_3.2}
This subsection outlines our experimental protocol, participant demographics, and methodological considerations.
\subsubsection{Experimental Protocol}\label{Sec_3.2.1}
At the beginning of the experiment, participants were required to sign a consent form for participation and use the collected hand movement frames for subsequent data analysis. The experimental procedure is described as follows.

\begin{figure}[t]
\centering
\vspace{-0.5em}\includegraphics[width=\linewidth]{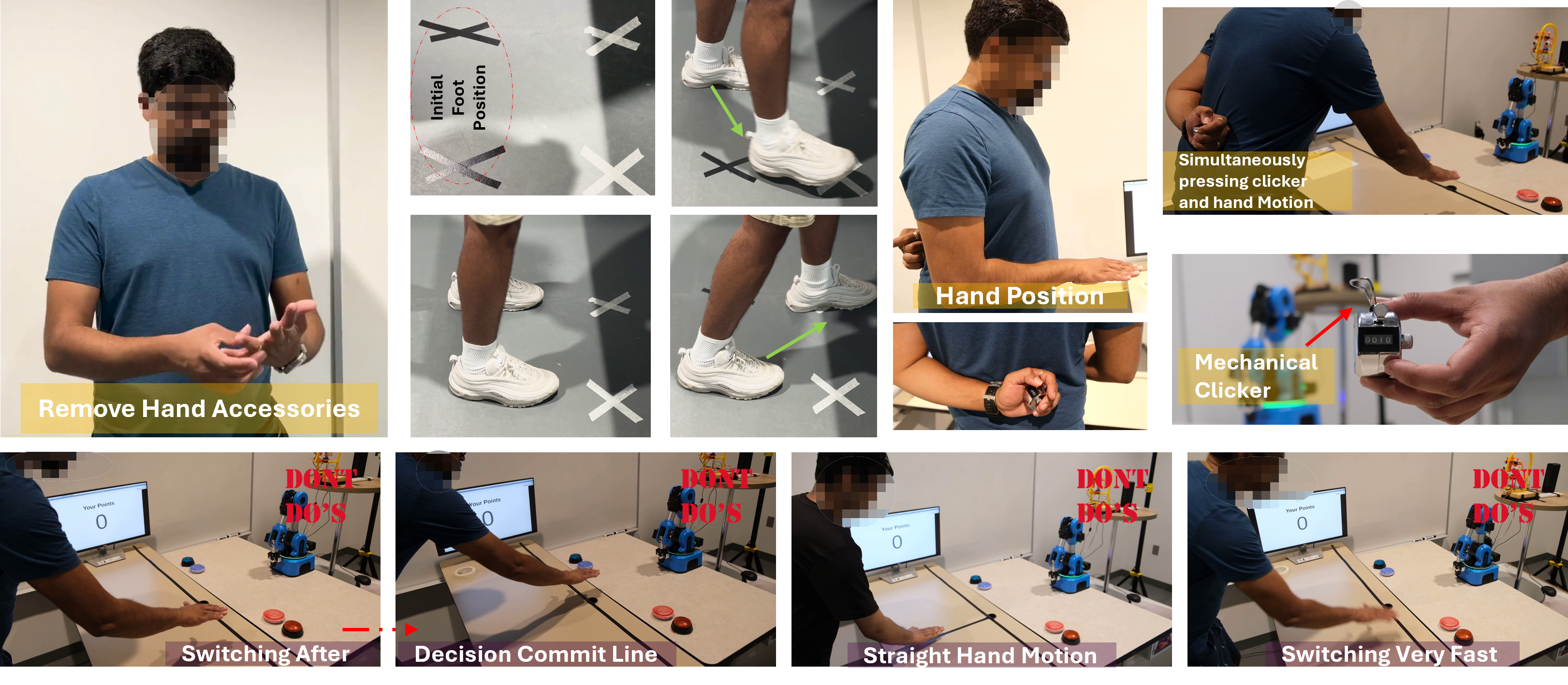}\vspace{-0.5em}
\caption{Pre-experiment instructions for participants, including steps for removing hand accessories, the initial standing position with allowed stepping marks, hand positioning, and the use of a mechanical clicker for cognitive load. Illustrations of 'Don't dos' such as switching after crossing the decision line, non-straight hand motion, and switching too quickly. (Human consent obtained for image use)}
\label{fig:2}\vspace{-0.5em}
\end{figure}

\vspace{0.5em}
\noindent \underline{\textbf{Step 1:}} Participants were instructed to remove any hand accessories and roll up their sleeves to ensure safety and unobstructed hand pose detection by the camera. Initially, they were directed to stand on two designated black X's marked on the floor, as illustrated in Fig.~\ref{fig:2}. Once the interaction began, participants were permitted to move to two designated white X's, enabling them to comfortably access the buzzer buttons.

\vspace{0.5em}
\noindent \textbf{\textit{Remark 1:}} Participants were not informed that their hand movements were being tracked, as the camera sensor was strategically positioned at a high elevation to remain unobtrusive.
\vspace{0.5em}

\noindent \underline{\textbf{Step 2:}} Following \underline{\textbf{Step 1}}, participants held a mechanical counter in their non-dominant hand behind their back while positioning their dominant hand forward, fingers extended toward the robotic arm in front of them.

\vspace{0.5em}
\noindent \underline{\textbf{Step 3:}} Upon hearing the auditory countdown “1, 2, 3, GO” from a speaker, each participant was instructed to immediately move their front-positioned hand to choose either the red or blue buzzer. Simultaneously, the robot initiated its own movement to select between the two options. \textit{The human participant’s objective was to press the same buzzer color as the robot at the end of each trial.}

\begin{definition}
We define \textit{human and robot opinions} as the nonlinearly varying arm path motions toward the red or blue option during the decision-making process. These hand motion trajectories reflect the dynamic intended preference as well as the steady-state choice, representing commitment to one of the options over time. Further details on the nonlinear opinion dynamics are provided in Section \ref{Sec_4}.
\end{definition}
\vspace{-0.5em}\noindent \textbf{\textit{Remark 2:}} To increase cognitive load, participants were simultaneously tasked with pressing the mechanical counter behind their backs exactly 10 times while moving toward the buzzer.

\vspace{0.5em}
\noindent \underline{\textbf{Step 4:}} During a trial, if participants press the same buzzer as the robot, they earn one point for the correct choice and an additional point for pressing the counter behind their back exactly 10 times. Failure to complete any of these tasks results in the loss of one point. Furthermore, violations of game rules specified in Remark 3 within a round will incur a one-point penalty.

\vspace{0.5em}
\noindent \underline{\textbf{Step 5:}} To match the robot's choice in real time, participants are allowed to switch their hand path once during each trial by observing the robot's behavior. Participants are completely unaware of the robot's future actions, and all human and robot opinions are formed in real time during the interaction.

\vspace{0.5em}
\noindent \underline{\textbf{Step 6:}} For each participant, this decision-making interaction game was repeated for eight iterations, with potential scores ranging from a minimum of -8 to a maximum of 16 points.

\noindent \textbf{\textit{Remark 3:}} Participants were not allowed to alter their hand path after crossing the decision commit line. They could switch paths only once per round and were required to maintain a nearly constant hand speed during the switch.

\vspace{0.5em}
\noindent \underline{\textbf{Step 7:}} In the first three of the eight trials, the robot's behavior was strategically configured to consistently disagree with the participant's choice, ensuring that it selected a different buzzer color. For instance, if the participant intended to choose red, the robot would continuously move towards blue, and vice versa. During these initial three trials, the robot's opinion was dynamically updated in a non-linear fashion, responding to participants hand movements to consistently choose the opposite option. 

\vspace{0.5em}
\noindent \textbf{\textit{Remark 4:}} Initially, robot behavior was designed to show disagreement to encourage human participants to increase their effort to collaborate and increase attentiveness in the game.

\vspace{0.5em}
\noindent \underline{\textbf{Step 8:}} During the \emph{first three trials}, \emph{both the robot and the human started without any external bias}, making decisions in real-time while the robotic eye remained in a neutral position. Beginning with the \emph{fourth trial}, the robotic eye was \emph{activated to introduce a visual bias toward a specific option}, thereby non-verbally communicating its intent and guiding the human participant toward pressing the same buzzer color, thus developing consensus. The robot, in response, \emph{dynamically updated its internal bias, overriding the social influence of the human participant's opinion and systematically aligning its decision-making with the option indicated by the robotic eye.}

\vspace{0.5em}
\noindent \textbf{\textit{Remark 5:}} From the outset of the experiment, participants were not provided any information about the robotic eye—its purpose, function, or even its significance. Following the initial three disagreement trials, the robotic eye was autonomously activated before the fourth trial, providing continuous visual gaze cues toward one of the options until the completion of that trial. Furthermore, from the fourth to the eighth trial, the intensity of the eye's gaze bias was progressively increased across different options to examine whether participants detected this gradual enhancement in visual cues, with the most pronounced and direct cue presented in the final, eighth trial.

At the beginning, the experimental coordinator verbally explained the entire procedure to each participant, supplemented by an instructional video illustrated in Fig.~\ref{fig:2}, outlining the Do's and Don'ts. Participants were instructed to maximize their score by selecting the same buzzer color as the robot, pressing the clicker exactly ten times, and adhering to the rules outlined in \textit{Remark 3}.

\subsubsection{Participants}\label{Sec_3.2.2}
A total of $51$ participants were recruited under IRB (Project Number: 2125317) through targeted outreach and direct engagement on the university campus, with minors excluded. The selection process prioritized diversity in educational background, age, and gender. To ensure the authenticity and reliability of the experimental results, individuals with prior knowledge of the experimental setup, components, or robot control algorithms, which could potentially bias results, were excluded. Participants ranged in age from $18$ to $55$ and represented diverse educational and professional backgrounds, including high school interns, post-doctorates, campus staff, Air Force veterans, and professional athletes, providing a broad range of perspectives. To ensure ethical practices, this experimental study does not involve any deceptive robot behaviors, and participants are debriefed afterward. There were no anticipated risks for participants, nor was there any financial compensation or prize for participation, ensuring the ethical conduct of the study.

\subsubsection{Methodological Considerations of Experimental Design}\label{Sec_3.2.3}
Our experiment adopts a \textbf{within-subjects design} wherein each participant experiences multiple conditions. Two \textbf{independent variables} vary across trials: (1) the robot’s eye gaze, which shifts from neutral (no visual communication during the first three of eight trials) to a pronounced “extreme” cue in the last round, and (2) the bias in robot’s opinion, transitioning from neutral to biased towards one of the option (red or blue) as the trials progress. The initial setup—with neutral eye gaze and unbiased opinion—functions as the \textbf{control condition}. We maintain \textbf{control variables} consistently throughout the study, including a fixed laboratory environment, a constant robot appearance, and identical instructions scripted to ensure the experimenter’s behavior is consistent across conditions. Additionally, while not directly manipulated, \textbf{co-variates} such as participant familiarity with robots (collected in a pre-experiment questionnaire) and individual differences in trust propensity may influence outcomes but remain outside our active manipulations.

To evaluate outcomes, we define two \textbf{dependent variables}. First, we measure \emph{consensus} by recording whether the human and robot select the same color buzzer (red or blue) during each trial. Second, to gauge \emph{trust} in the robot’s visual cues once the gaze becomes active, we examine how directly (without any alterations in their hand path) and rapidly participants press the buzzer option communicated by the robot’s eye gaze. We collect \textbf{quantitative data}, including each participant’s buzzer choice and any deviations in hand movement, which are automatically captured through the overhead camera sensor and the human opinion observation algorithm described in Subsection \ref{Sec_4.5}. To gain a deeper understanding of the decision-making process, we also collect \textbf{qualitative data} via post-experiment interviews and open-ended survey questions. These address whether participants noticed and followed the robot’s cues, why (or why not) they were persuaded by them, and invite feedback on improving the study. By combining controlled manipulations of robot gaze and bias, consistent experimental conditions, and a blend of objective and subjective measures, this design aims to clarify how evolving robot communication strategies influence human-robot collaboration and consensus building.

\vspace{0.5em}
\noindent \textbf{\textit{Research Questions and Hypotheses:}} To evaluate the following research questions:  
\begin{itemize}
   \item \textbf{RQ1:} Can non-verbal communication through robotic eye-gaze combined with biased robot opinion establish consensus in human-robot co-learning?  
   \item \textbf{RQ2:} Can participants' trust in robotic guidance develop progressively through increased exposure to consistent robotic non-verbal cues?  
\end{itemize}
\noindent We formulate the following alternative hypotheses:  
\begin{itemize}
   \item \textbf{H1:} The combination of robotic eye-gaze cues and biased robot opinion significantly increases consensus rates in human-robot co-learning environments compared to no-cue conditions.
   \item \textbf{H2:} Participants' trust in robotic guidance increases progressively with repeated exposure to consistent robotic non-verbal cues, as measured by decreasing hesitation behaviors and increasing consensus rates.
\end{itemize} 
\noindent The corresponding \textit{null hypotheses} assume no significant effect of robotic communication cues on consensus achievement or progressive trust development.

\section{Nonlinear Opinion Dynamics for Robot Behavioral Control}\label{Sec_4}

To design the robot's behavior for non-verbal communication of its intended choice, we first conducted a theoretical analysis of the dynamic opinion formation between the human and the robot during co-learning process. 

\subsection{Mathematical Preliminaries and Model Formulation}\label{Sec_4.1}
We model the human participant and the robot as two interacting agents. Each agent has the option of pressing a red or blue buzzer, forming their continuous opinion $z_i \!\in\! \mathbb{R}$ with $i\!\in\! \{r, h\}$, where $i\!=\!r$ represents the robot and $h$ represents the human participant. At any given moment during the interaction, if the opinion variable \( z_i \) satisfies that \( z_i \!>\! 0 \), agent \( i \) is predisposed to press the red buzzer; if \( z_i \!<\! 0 \), the blue buzzer is the intended choice. The magnitude \( |z_i| \) quantifies the strength of the agent's conviction towards pressing either the red or blue buzzer. 

Following the model presented by equation~(14) in \cite{Bizyaeva2023}, and incorporating the saturation function \( \hat{S}_1(y)\! = \!\tanh(y) \), we define the option dynamics of the robot and human as \( \dot{z}_r \) and \( \dot{z}_h \), respectively, and model their interaction as a two-agent, two-option opinion network as 
\begin{align}
\dot{z}_r &= -d_r z_r + u_r \tanh(\alpha_r z_r + \gamma_r a_{rh} z_h) + b_r, \label{eq:1}\\
\dot{z}_h &= -d_h z_h + u_h \tanh(\alpha_h z_h + \gamma_h a_{hr} z_r) + b_h,  \label{eq:2}
\end{align}
where \(d_r \!>\! 0\) and \(d_h \!>\! 0\) are decay constants representing the attenuation of opinions over time, \(u_r\) and \(u_h\) denote quantitative social influence parameters, and \(\alpha_r\) and \(\alpha_h\) are weights for self-opinion reinforcement. Parameters \(\gamma_r\) and \(\gamma_h\) represent inter-agent gains of influence on the same opinion, while \(b_r\) and \(b_h\) act as bias or external stimulus inputs for the robot and human participant, respectively. 

Let a matrix \( A \) denote the unweighted adjacency matrix of the two agents network, with $ A\!=\![0, a_{rh}; a_{hr}, 0]$, where the off-diagonal elements \( a_{rh} \) and \( a_{hr} \) represent binary directed communication edges from the robot to the human and from the human to the robot, respectively. These elements in \( A \) are set to \( 1 \) if both the robot and human continuously observe each other, and \( 0 \) otherwise. For our experiments, the robot and human interact in a continuous closed-loop during decision making. Accordingly, we set \( a_{rh} \!=\! a_{hr} \!= \!1 \), resulting in an unweighted adjacency matrix \( A \) with eigenvalues \( \lambda_{\max} \!= \!1 \) and \( \lambda_{\min}\! =\! -\!1 \). We define \textbf{consensus} as the condition \( \text{sign}(z_r)\! =\! \text{sign}(z_h) \), where both the human and the robot select the same buzzer color at steady state or at the end of a trial. Conversely, \textbf{dissensus} occurs when \( \text{sign}(z_r) \!\neq\! \text{sign}(z_h) \), indicating that the human and robot choose different buzzer colors. Depending on their respective opinions, either state may emerge during the course of interaction.


\subsection{Design of Dissensus Behavior for Robot in Trials 1\textsuperscript{st} to 3\textsuperscript{rd}}\label{Sec_4.3}

By considering a homogeneous opinion system with identical human nd robot opinion dynamics parameters, and setting \( b_r \approx 0 \) and \( b_h \approx 0 \) (as studied in~\cite{Leonard2024}), a Jacobian matrix \( J \) for \eqref{eq:1} and \eqref{eq:2} can be derived as follows:
\begin{equation}
J = \begin{bmatrix}
    -d + u \alpha \left(\sech(\alpha z_1 + \gamma a_{12} z_2)\right)^2 & u \gamma \left(\sech(\alpha z_1 + \gamma a_{12} z_2)\right)^2 \\
    u \gamma \left(\sech(\alpha z_2 + \gamma a_{21} z_1)\right)^2 & -d + u \alpha \left(\sech(\alpha z_2 + \gamma a_{21} z_1)\right)^2
\end{bmatrix}.
\label{eq:3}
\end{equation}
Evaluating the Jacobian \( J \) from (\ref{eq:3}) at the equilibrium \( [z_r, z_h]^\top \!=\! [0, 0]^\top \) results in \( J \!=\! (-d \!+\! u\alpha)I \!+\! u\gamma A \). Substituting \( J \) into its eigenvalue equation \( Jv \!=\! \lambda_J v \) and rearranging, we obtain the eigenvalue of \( A \) as \( \lambda \!=\! \frac{\lambda_J + d - u\alpha}{u} \). Now, the key results required for decision-making and designing dissensus behavior between two agents, derived from \cite{Bizyaeva2023, Leonard2024}, are:

\noindent 1. A pitchfork bifurcation occurs at critical attention point $ u \!=\! u^* \!=\! \frac{d}{\alpha + \gamma \lambda} $ where $\lambda_J \!=\! 0$ and for $ u \!>\! u^* $ and $\lambda_J \!>\! 0$, the system transitions from a neutral unopinionated stable state to an opinionated unstable state, along with two stable equilibria. For \( u \!<\! u^* \) (where \( \lambda_J \!<\! 0 \)), opinions remain undecided.\\
2. Setting $ u \!>\! u_d^*\!=\! \frac{d}{\alpha + \gamma \lambda_{\min}} $, $\lambda_{\min} \!=\! \min(\lambda)$ and $\gamma \!<\! 0$, the opinion network is guaranteed to sustain a state of dissensus between the agents as a unique stable equilibrium.

As outlined in \underline{\textbf{Step 7}} of the experimental procedure in Section \ref{Sec_3.2.1}, the robot’s behavior was deliberately configured to induce simulated disagreement during the first three of eight trials by consistently selecting the buzzer color opposite to the participant’s choice. This was achieved by utilizing the above key results by setting \( u \!>\! u_d^* \), \( \lambda_{\min} \!=\! -1 \), and \( \gamma \!<\! 0 \), with both the human and robot bias parameters set to \( b_r \!=\! b_h \!=\! 0 \). Before detailing the design of robot behavior for consensus and the use of bias to provide visual cues influencing human opinion, we briefly refocus the reader’s attention on the overarching objective of this paper. By relaxing the small-bias assumption \cite{Bizyaeva2023, Leonard2024}, we explore dynamic bias adjustment within nonlinear opinion dynamics for practical implementation in human-robot interaction. For subsequent trials, beginning with the fourth, the experimental protocol was designed to facilitate a transition from dissensus to consensus. Under the sustained conditions of $u \!>\! u_d^*$ and $\gamma \!<\! 0$, which promote disagreement between human and robot, the theoretical analysis in Section \ref{Sec_4} will explain the mechanisms that can lead to consensus, ultimately enabling coordinated actions, such as the selection of the same buzzer color by both agents. 


\subsection{Design and Analysis for Bias Controlled Transition to Consensus}\label{Sec_4.4}

For fully controlled and practically applicable consensus transitions, we propose leveraging the bias parameters \( b_r \) and \( b_h \). Here, \( b_h \) represents the external bias of human guided by the non-verbal visual cues from the robot's eyes, influencing human opinion formation. In contrast, \( b_r \) denotes the robot's bias, which we have complete control over and thus we developed a dynamic updating rules for $b_r$. Both \( b_r \) and \( b_h \) are used to modulate the human and robot opinions \( z_r \) and \( z_h \), respectively, enabling controlled transitions from dissensus to consensus between the human and the robot. Additionally, while we can observe human opinions during interaction, we assume the non-linear opinion dynamics in \eqref{eq:2} reflect internal processes in the human brain, limiting our ability to fully control human opinion formation. This is discussed in detail and illustrated in Fig.~\ref{fig:4}(b) within Section \ref{Sec_4.5}. 

To design a method for dynamically updating the robot bias \( b_r \) and utilizing it as a control parameter for consensus transitions, we conducted numerical simulations, as presented in Section \ref{Sec_4.4.1}. These simulations included parameter sweeps over \( b_r \) and \( b_h \) and equilibrium nullcline analyses of the opinion dynamics \eqref{eq:1} and \eqref{eq:2} to examine how human and robot opinion dynamics evolve when \( b_r \) and \( b_h \) are varied dynamically, rather than being fixed at zero.

\subsubsection{Parameter Sweep and Nullcline Analyses for Biases}\label{Sec_4.4.1}
\begin{figure}[t]
    \centering
    \includegraphics[width=\linewidth]{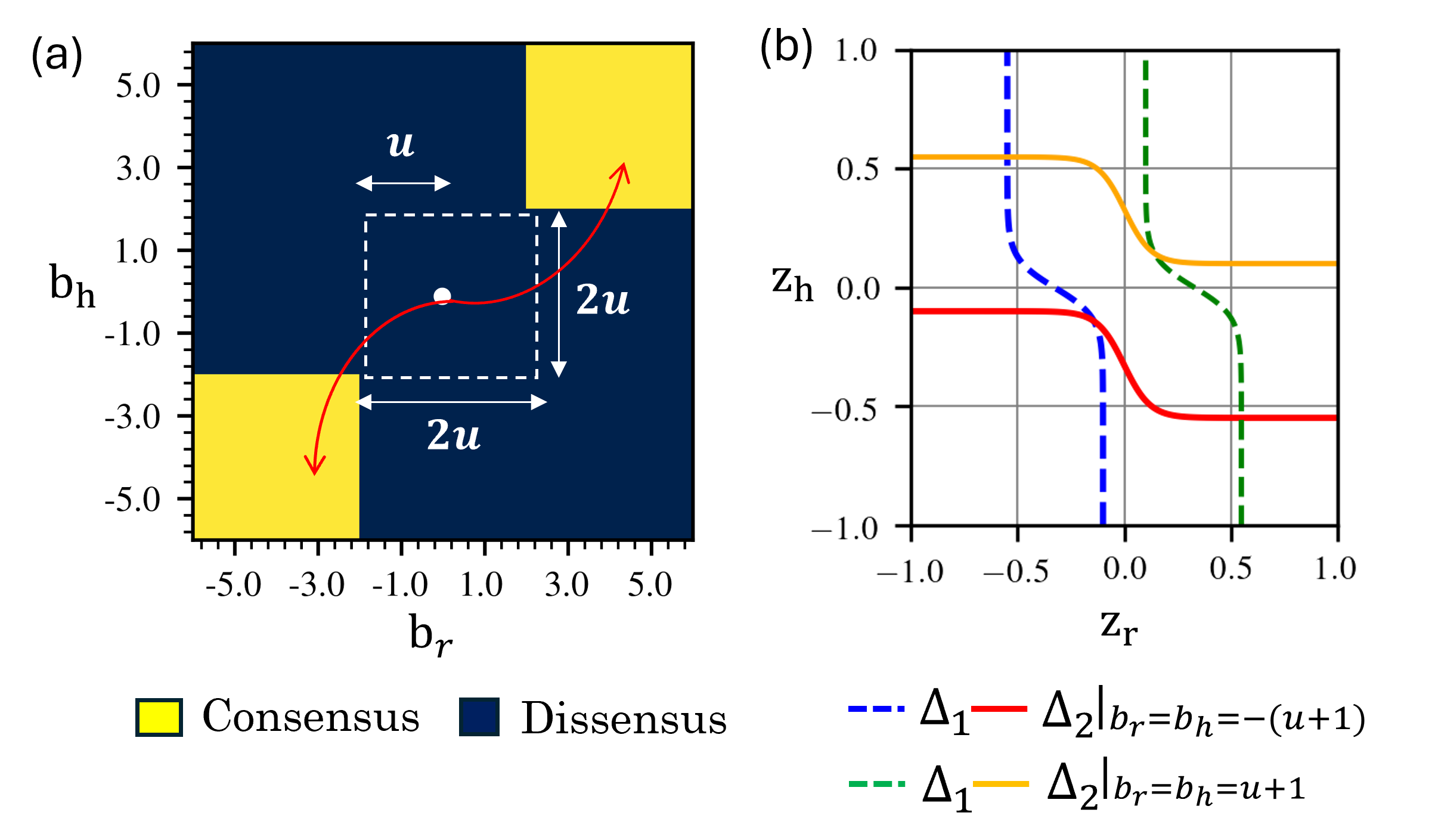}\vspace{-1em}
    \caption{(a) Parameter sweep over $\left(b_r, b_h\right)$, highlighting consensus (yellow) and dissensus (blue) regions. (b) Zero-level contours of $\Delta_1, \Delta_2$ in the $\left(z_r, z_h\right)$ plane for $b_r=b_h= \pm(u+1)$ illustrate the number of equilibrium solutions.}
    \label{fig:3}
\end{figure}

Parameter sweeps of the biases \( b_r \) and \( b_h \) within the range \([-6, 6]\) were conducted using the optimized parameter set under the pre-configured conditions \( u \!>\! u^*\! =\! d/({\alpha \!+\! \gamma \lambda_{\min}}) \) and \( \gamma \!<\! 0 \), ensuring a dissensus regime. As can be observed from Fig.~\ref{fig:3}(a), achieving consensus requires the bias pair \( (b_r, b_h) \) to exceed the boundaries of a square (indicated by a white dashed line) centered at \( (0, 0) \) with side length \( 2u \). Furthermore, once outside this square region, both biases must share the same sign, i.e., \( \sign(b_h) \!=\! \sign(b_r) \), where \( \sign(x) \!=\! -\!1 \) if \( x \!<\! 0 \), \( \sign(x) \!=\! 0 \) if \( x \!=\! 0 \), and \( \sign(x) \!=\! 1 \) if \( x \!>\! 0 \). And consensus between the agents results in both pressing the red buzzer when \( \sign(b_r) \!=\! \sign(b_h) \!=\! 1 \), whereas \( \sign(b_r)\! =\! \sign(b_h) \!=\! -\!1 \) leads to both selecting the blue buzzer.

To determine the number and stability of equilibrium solutions for human and robot opinions once the biases \( b_r \) and \( b_h \) exceed the social attention threshold \( u \), we perform a nullcline analysis. Intuitively, the number of equilibria indicates how many distinct long-term outcomes are possible—whether the agents converge to a single stable consensus or diverge. Once consensus is reached through dynamic bias adjustment, nullcline analysis helps assess whether the agents will remain committed to the agreed consensus option or whether the system permits divergence to alternative opinions over time. At equilibrium, where \( b_r \!\neq\! 0 \) and \( b_h \!\neq\! 0 \), rearranging \eqref{eq:1} and \eqref{eq:2} yields:

\vspace{-0.5cm} 
\begin{align}
\Delta_1 \big|_{b_r \neq 0, b_h \neq 0} (z_r, z_h) = \tanh(\alpha z_r + \gamma z_h) - \frac{d z_r - b_r}{u}, \label{eq:4}\\
\Delta_2 \big|_{b_r \neq 0, b_h \neq 0} (z_r, z_h) = \tanh(\alpha z_h + \gamma z_r) - \frac{d z_h - b_h}{u}.\label{eq:5}
\end{align}

By plotting the zero-level contours of \( \Delta_1(z_r, z_h) \) and \( \Delta_2(z_r, z_h) \) in the \( (z_r, z_h) \) plane, we identify the loci where each equation is satisfied independently. The intersection points of these contours represent the equilibrium solutions of the system, as they satisfy both \( \Delta_1 \!=\! 0 \) and \( \Delta_2 \!=\! 0 \). The number of such intersections indicates the number of equilibrium solutions. As shown in Fig.~\ref{fig:3}(b), the zero-level contours of \( \Delta_1(z_r, z_h) \) and \( \Delta_2(z_r, z_h) \) are plotted under the equilibrium condition \( \dot{z}_r\!=\! \dot{z}_h \!=\! 0 \) for two bias settings: \( b_r \!=\! b_h \!=\! -(u\!+\!1) \) (Blue and Red) and \( b_r \!=\! b_h \!=\! u\!+\!1 \) (Orange and Green). In both cases, a single unique intersection is observed, indicating one stable equilibrium. This stability is further confirmed by the negative eigenvalues of the Jacobian matrix evaluated at the equilibrium point. From these observations, we conclude that for our chosen parameter set, if the magnitudes of \( b_r \) and \( b_h \) exceed the social attention parameter \( u \) and both biases share the same sign, the system settles into a single, stable consensus equilibrium. This suggests that once the human’s and robot’s biases surpass their social attention threshold and align on a single option (red or blue), they remain committed to that choice, indicating stability toward one solution with no drift toward the alternative.

\noindent\textbf{Note:} The claim of a single unique equilibrium presented above is valid for the chosen optimized parameters specifically for our human-robot interaction experimental setup. However, the bias parameters can exhibit multi-stable, multi-phase equilibrium solutions. A detailed analysis of such opinion dynamics behavior is beyond the scope of this paper.

\subsubsection{Proposed Dynamic Bias Updating Rule for Robot}\label{Sec_4.4.2}
Summarizing insights from the numerical analyses above, we can state that within our experimental framework—where a human participant and a robot act as two agents choosing between a red and blue buzzer, for the disagreement trials (1\textsuperscript{st} to 3\textsuperscript{rd})—a transition from dissensus to consensus in their opinions (\(z_r, z_h\)) can only occur if the following conditions on both human and robot biases are met:
\begin{enumerate}
    \item The biases of the human and robot, \(b_r\) and \(b_h\), must have the same sign, i.e., \(\operatorname{sign}(b_r) = \operatorname{sign}(b_h)\).

    \item The magnitudes of these biases, \(|b_r|\) and \(|b_h|\), must exceed their respective critical bias (attention) thresholds \(b_r^*\) and \(b_h^*\), defined as \(b_i^* = u_i^* > \frac{d_i}{\alpha_i + \gamma_i \lambda_{\min}} > 0\) for \(i \in \{r, h\}\), i.e., \(|b_r| > b_r^*\) and \(|b_h| > b_h^*\). Specifically:
    \begin{itemize}
        \item For agreement on the red buzzer (\(\text{sign}(z_r), \text{sign}(z_h) \in \{+\}\)), both biases \(b_r\) and \(b_h\) must be positive, \(\text{sign}(b_r) = \text{sign}(b_h) = +1\).
        
        \item For agreement on the blue buzzer (\(\text{sign}(z_r), \text{sign}(z_h) \in \{-\}\)), both biases \(b_r\) and \(b_h\) must be negative, \(\text{sign}(b_r) = \text{sign}(b_h) = -1\).
    \end{itemize}
\end{enumerate}

To achieve this transition to consensus from dissensus under stated bias conditions, we present dynamic biases model for robot bias $\dot{b_r}$ as
\begin{equation}
\dot{b}_r = \sigma \cdot z_r \cdot \text{sgn}(z_h) + \beta \cdot \max(0, b^*_r - |b_r|)),
\label{eq:6}
\end{equation}
The first term \( \sigma \cdot z_r \cdot \text{sgn}(z_h) \) in the bias updating rule aligns the signs of \( b_r \) and \( b_h \) according to the selected agreement option (pressing red or blue), effectively merging two perspectives and reducing potential conflicts by leveraging the interaction of the opinions \( z_r \) and \( z_h \). And since the robot can observe the action taken by the human, the $\text{sgn}(z_h)$ can be directly measured even though the robot cannot directly access the value of $z_h$. The second term, \( \beta \cdot \max(0, b_r^* \!-\! |b_r|) \), elevates the bias \( b_r \) above the designated threshold \( b_r^* \), defined such that \( b_r^* \!=\! u_r^* \!>\! \frac{d_r}{\alpha_r + \gamma_r \lambda_{\min}} \!>\! 0 \), ensuring that the robot bias exceeds the values required for stable agreement dynamics. The proposed bias dynamics model draws on real-world intuition, showing that the inclination to favor a specific option (red or blue) intensifies when \( b_r \) surpasses the maximum impact of collective social influence exerted by the human counterpart, quantified as \( u_r \tanh(\alpha_r z_r \!+\! \gamma_r z_h) \), where the maximum value of \( \tanh(\alpha_r z_r \!+\! \gamma_r z_h) \) is one. Consequently, when \( b_r \) exceeds the attention threshold \( u_r \), the robot’s opinion dynamics progressively align with the selected biased option, reflecting a deeper integration of individual and collective preferences.
\begin{figure}[ht]
\centering
\includegraphics[width=\linewidth]{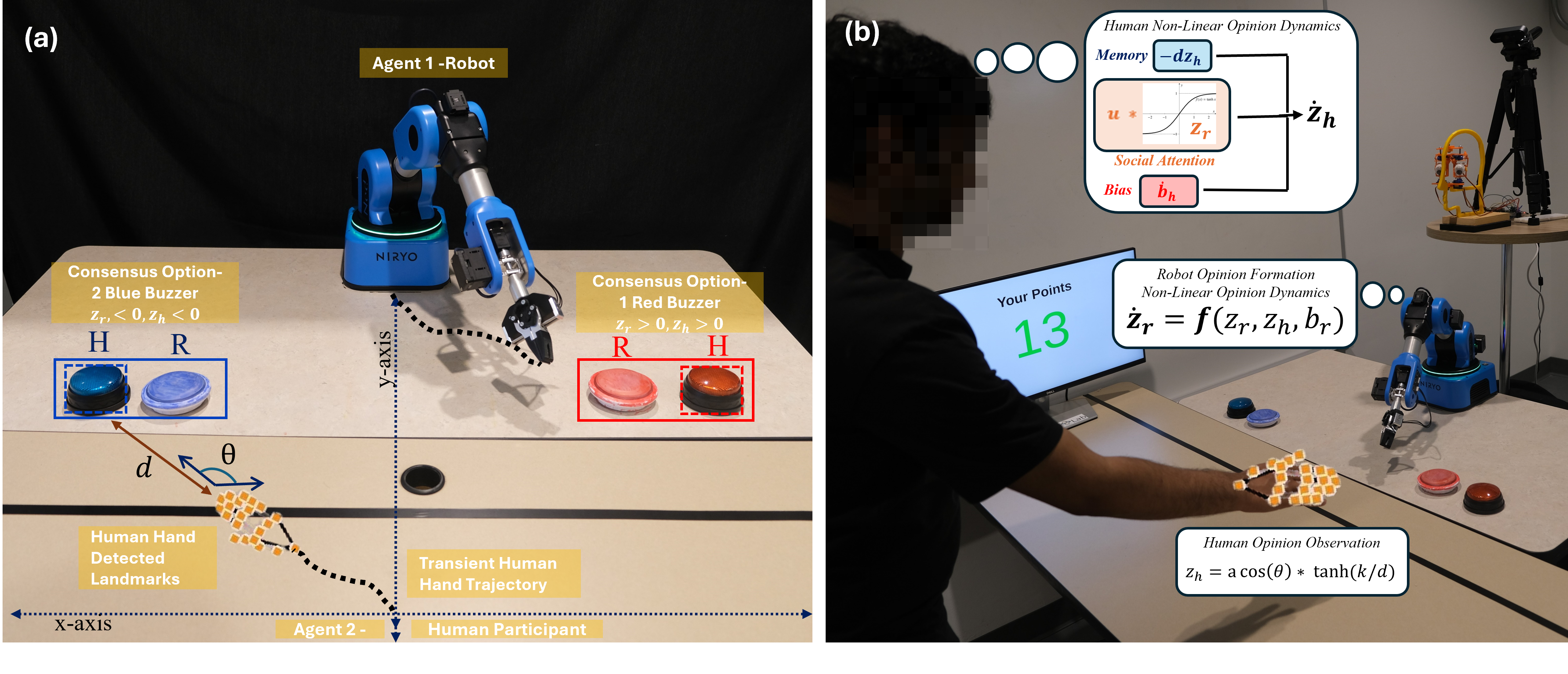}\vspace{-1em}
\caption{(a) Illustration of the calculation of the angle \( \theta \) and distance \( d \) for human opinion observation, showing how the robot captures and interprets the human's decision-making process. (b) Illustration of a comprehensive mental model integrating human internal opinion dynamics, observed opinion dynamics, and the robot's non-linear opinion dynamics, demonstrating how the decision-making process is anticipated during interaction.}\vspace{-1em}
\label{fig:4}
\end{figure}

\noindent\textbf{Note:} Numerical analyses in Section \ref{Sec_4.4.1} indicate that, to achieve consensus, the human’s bias \(b_h\) must also be updated alongside the robot’s bias \(b_r\). However, we cannot directly control \(b_h\) through a mathematical equation. Thus, starting from the 4\textsuperscript{th} trial, we employ robotic eye gaze to visually convey the robot’s biased opinion (i.e., its favored option). By influencing the human’s bias via these gaze cues, we anticipate that human participants will ultimately select the same buzzer option as the robot, thereby achieving consensus. Furthermore, if the experimental data show that humans do indeed follow the robot’s visual clues, it will confirm that the human’s opinion is actively shifted (i.e., ``biased'') to match the robot’s stance, surpassing any maximum social attention and thus validating our theoretical approach experimentally for biasing human behavior toward consensus.

\vspace{0.5em}
\noindent \textit{\textbf{Remark 7:}} We need the dynamic model for \(b_r\), rather than simply configuring the initial settings by assigning static \(b_r \!=\! |u| \!+\! b_r\), as such settings would place the opinion system in agreement from the outset of the formation of the opinions. Also, these static bias configurations contradict the primary goal of this research, which is to allow dynamic bias to evolve over time when a disagreement between agents is detected, leading the opinions to reach a consensus naturally. Furthermore, human biases are not static in the real world but form, evolve, and dissipate over time based on accumulating evidence \cite{Park2022, Zylberberg2018}. This underscores the need for a dynamic bias model presented in \eqref{eq:6} which can be used as dynamic control to drive the system toward consensus. 
 
\subsection{Human Opinion Observation and Robot Control Algorithm}\label{Sec_4.5}

Although it is anticipated that human internal decision-making adheres to non-linear opinion dynamics given in \eqref{eq:2}, which effectively incorporate elements such as memory retention of previous choices, non-linear social attention to robotic inputs, and the influence of external biases, however, direct control over human opinion is not feasible. Only observation of human opinion is possible during interactive settings, as Fig.~\ref{fig:4}(b) illustrates. Consequently, to facilitate the empirical observation of human opinions during these interactions, we propose 
\vspace{-2mm}
\begin{equation}
\hat{z}_h = a \cdot \cos(\theta) \cdot \tanh\left({k}/{d}\right)
\label{eq:7}
\vspace{-2mm}
\end{equation}

where \( \hat{z}_h \) quantifies the observed human opinion. The term \( a \cdot \cos(\theta) \) calculates the directional intent of the human opinion towards a specific option, where \( \theta \) is the angle between the human movement vector and the x-axis as shown in Fig.~\ref{fig:4}(a). Additionally, the hyperbolic tangent function, \( \tanh\left(\frac{k}{d}\right) \), modulates the intensity of the opinion based on the normalized distance \( d \) from the human hand to the target buzzers. The parameters \( k \) and \( a \) are scaling weights that adjust the influence of distance and directional intent, respectively.

Algorithm \ref{algorithm_1} outlines the process of observing human opinions and forming corresponding robotic opinions during interaction. It takes real-time human hand motion as input, captured by an overhead camera sensor, and a predefined convergence option \( O \) (either red or blue), which both the human and robot are expected to agree upon at steady state. From participant's real-time captured hand motion frames, we utilized the state-of-the-art deep learning-based hand-tracking vision module, which continuously detects the complete set of 2D pose coordinates \(\{(x_i, y_i)\}_{i=1}^N\) of all fingertips, knuckles, and thumbs. The twenty-one detected 2D coordinates are averaged to derive a single central pose, representing the dynamic movement of the hand throughout the participant's interaction with the robot and decision-making process. Additionally, the vision tracking algorithm establishes two bounding boxes around the options—the blue and red buzzers—which Algorithm \ref{algorithm_1} uses as fixed positional input data.

The human hand landmarks and the formed robotic opinion \( z_r \) are transmitted to Algorithm \ref{algorithm_2}, which then controls robot actions. Furthermore, Algorithm \ref{algorithm_1} calculates the norm of the human hand movement vector \(\|\mathbf{M}\|\) using the current position vector \(\mathbf{p}(t)\) and the immediately preceding position vector \(\mathbf{p}_{-1}(t)\). Utilizing the movement vector \( \mathbf{M} \), \( \theta \) is computed and normalized to restrict its range to the first and second quadrants according to Fig.~\ref{fig:4}(a), aligning the calculation of opinions with the choice of either the red or blue buzzer. If \( \theta \) lies within \([ \frac{\pi}{2}, \pi )\), \( d \) quantifies the distance to the blue buzzer, indicating a directional intent towards it. Conversely, when \( \theta \) is within \([0, \frac{\pi}{2})\), \( d \) measures proximity to the red buzzer, thus determining the targeted buzzer based on hand orientation. Across trials 1\textsuperscript{st} to 8\textsuperscript{th}, the observed human opinion \( \hat{z}_h \) and robot opinion \( z_r \) are determined, with the robot's non-linear opinion dynamics configured to \( u_r > u_d^* = \frac{d}{\alpha + \gamma \lambda_{\min}} \), where \( \gamma < 0 \) intentionally promotes disagreement with human opinions. However, from the $4^{\text{th}}$ to the $8^{\text{th}}$ trial, the \textbf{Bias Consensus Algorithm} is activated, dynamically adjusting the robot's opinion \( z_r \) and bias \( b_r \) based on \eqref{eq:6} toward the initially inputted consensus option \( O \) in Algorithm \ref{algorithm_1}. The utilized parameters for the dynamic bias model \( \dot{b_r} \), along with the opinion dynamics \( \dot{z_r} \) and \( \hat{z}_h \), are discussed in Section \ref{Sec_4.6}.

\setlength{\algomargin}{0em}  
\begin{algorithm}[H]\label{algorithm_1}
\footnotesize  
\caption{Human Opinion Observation and Robot Opinion Formation Algorithm}
\DontPrintSemicolon 
\KwIn{\( O \in \{\text{``red"}, \text{``blue"}\} \), \text{Camera feed of human hand motion}}
\tcp*{Executed once per trial. \(O\) is the consensus option for trials 4\textsuperscript{th}–8\textsuperscript{th}; camera stream is used per-frame within trial.}
\KwOut{\textcolor{Red}{\textbf{Robot Action}} in Algorithm \ref{algorithm_2}}
\BlankLine
\KwData{\( \mathbf{p}_{\text{buzzer}} = \left\{ \text{``blue"}: [(x_{1b}, y_{1b}), (x_{2b}, y_{2b})], \text{``red"}: [(x_{1r}, y_{1r}), (x_{2r}, y_{2r})] \right\} \)}
\tcp*{Position vector of buzzers bounding boxes relative to hand pose}
\KwTrigger{Decision-making begins at the cue: “1, 2, 3... go.”}

\While{\textbf{true}}{
    \( I \gets \text{Capture Hand Frames} \),  
    \( \{(x_i, y_i)\}_{i=1}^N \gets \text{Detect 2D Hand Landmarks}(I) \) \\
    \( \mathbf{p}(t) = \left[\frac{1}{N} \sum_{i=1}^N x_i, \frac{1}{N} \sum_{i=1}^N y_i\right]^\top \) \tcp*{Compute current hand position vector}
    \( \mathbf{p}(t-1) = \left[\frac{1}{N} \sum_{i=1}^N x_{i, t-1}, \frac{1}{N} \sum_{i=1}^N y_{i, t-1}\right]^\top \) \tcp*{Previous hand position vector}
    \( \mathbf{M} = \mathbf{p}(t) - \mathbf{p}(t-1) \), \( \|\mathbf{M}\| = \sqrt{\mathbf{M}^\top \mathbf{M}} \) \tcp*{Human hand movement vector and norm}
    \( \theta = \arctan2\left(-\mathbf{M}_y, \mathbf{M}_x\right) \) \tcp*{Hand movement angle relative to x-axis.}
    \( \theta = \theta - 2\pi \cdot \left\lfloor \frac{\theta}{2\pi} \right\rfloor \) \tcp*{Adjust \(\theta\) to be non-negative and less than \(2\pi\)}
    \uIf{\( \frac{3\pi}{2} \leq \theta < 2\pi \)}{
        \( \theta = \pi - (2\pi - \theta) \) \tcp*{Mirror \(\theta\) to second quadrant}
    }
    \uElseIf{\( \pi \leq \theta < \frac{3\pi}{2} \)}{
        \( \theta = \theta - \pi \) \tcp*{Shift \(\theta\) to first quadrant}
    }
    \uIf{\(\frac{\pi}{2} \leq \theta < \pi\)}{
        \( d = \| \mathbf{p}(t) - \mathbf{p}_{\text{buzzer}}[\text{``blue"}][0] \| \) \tcp*{Distance of hand pose to blue} 
    }
    \uElseIf{\(0 \leq \theta < \frac{\pi}{2}\)}{
        \( d = \| \mathbf{p}(t) - \mathbf{p}_{\text{buzzer}}[\text{``red"}][0] \| \) \tcp*{Distance of hand pose to red} 
    }
    \( \hat{z}_h = a \cdot \cos(\theta) \cdot \tanh(k / d) \) \tcp*{Observe human opinion \( \hat{z}_h \) based on \( d \) and \(\theta\)}
    \( \dot{z}_r = -d_r z_r + u_r \tanh(\alpha_r z_r + \gamma_r \hat{z}_h) + b_r \) \tcp*{Formulates robot opinion \( z_r \) under persistent disagreement conditions where \( u_r > u_d^* = \frac{d}{\alpha + \gamma \lambda_{\min}} \), and, \( \gamma < 0 \).}
\textbf{Bias Consensus Algorithm (Activated for only trial 4\textsuperscript{th} to 8\textsuperscript{th})}\;
    \While{\( z_r \cdot \hat{z}_h < 0 \)}{
         \lIf{$O = \text{``Red"}$}{$\sigma \gets -1$}
         \lElse{$\sigma \gets +1$}
        \( \beta = -K \cdot \sigma \) \tcp*{Define \( \beta \) in terms of \( K \) and \( \sigma \)}
        \( \dot{b}_r = \sigma (z_r \cdot \text{sgn}(\hat{z}_h) - K \cdot \max(0, b_r^* - \lvert b_r \rvert)) \) \tcp*{Adjust \( b_r \) dynamics}
        \( \dot{z}_r = -d_r z_r + u_r \tanh(\alpha_r z_r + \gamma_r \hat{z}_h) + b_r \) \tcp*{Update \( z_r \) dynamics}
    }
    \textbf{On Exit} (When $z_r \cdot \hat{z}_h > 0$) \tcp*{Consensus achieved}
    \( \dot{b}_r \gets 0 \) \tcp*{Stop updating \( b_r \)}
}
\Transmit{\( z_r \), \( \{(x_i, y_i)\}_{i=1}^N \), and \( \mathbf{p}_{\text{buzzer}} \) to Algorithm \ref{algorithm_2}.} 
\end{algorithm}

\setlength{\algomargin}{0em}  
\begin{algorithm}\label{algorithm_2}
\footnotesize  
\caption{Robot Behaviour Control Algorithm}
\DontPrintSemicolon 
\textbf{Receive:} \( z_r \), \( \{(x_i, y_i)\}_{i=1}^N \), and \( \mathbf{p}_{\text{buzzer}} \) from Algorithm \ref{algorithm_1}. \\
\( \sigma_{\text{sgn}(z_r)} \gets 0, \sigma_{\text{sgn}(z_r)} \in \{-1, +1, 0\} \) \tcp*{\( \sigma_{\text{sgn}(z_r)} \) is preceding value of \(\text{sgn}(z_r)\)}
\While{\textbf{true}}{
    \uIf{\( \sigma_{\text{sgn}(z_r)} = 0 \)}{    
        \uIf{\( \text{sgn}(z_r) = +1 \)}{ 
            \textcolor{Red}{\textbf{Robot Action:}} Move Towards Red \tcp*{Move through first value of \(\text{sgn}(z_r)\)} 
        }
        \Else{
            \textcolor{Red}{\textbf{Robot Action:}} Move Towards Blue 
        }
        \( \sigma_{\text{sgn}(z_r)} \gets \text{sgn}(z_r) \) \tcp*{Update sign for the next cycle}
    }
    \If{\( \sigma_{\text{sgn}(z_r)} \neq 0 \) \textbf{and} \( \sigma_{\text{sgn}(z_r)} \neq \text{sgn}(z_r) \)}{
        \( C \gets C + 1 \)  \tcp*{Increment change counter on \( z_r \) sign change}
        \( \vec{T}[0] \gets \text{True} \) \tcp*{Mark the first transition state}
    }
    \If{\( \sigma_{\text{sgn}(z_r)} \neq 0 \)}{
        \For{\( i \gets 0 \) \textbf{to} 1}{
            \If{\( \vec{T}[i] \) \textbf{and} \( \sigma_{\text{sgn}(z_r)} = \text{sgn}(z_r) \)}{
                \( \vec{T}[i+1] \gets \text{True} \) \;
                \( \vec{T}[i] \gets \text{False} \) \;
                \( C \gets C + 1 \) \tcp*{Update state and increment \( C \)}
            }
        }
        \If{\( \vec{T}[2] \) \textbf{and} \( \sigma_{\text{sgn}(z_r)} = \text{sgn}(z_r) \)}{
            \( C \gets C + 1 \) \Comment{Confirm final state and increment counter}
            \If{\( C \geq C_{\text{max}} \)}{
                \textcolor{Red}{\textbf{Robot Action:}} Stop Movement \tcp*{Pause 0.1s before switch}
                \( C \gets 0 \) \tcp*{Reset change counter for next change}
                \( \vec{T} \gets [False, False, False] \) \tcp*{Reset transition states}
            }
        }
    }
    \If{\(\sigma_{\text{sgn}(z_r)} \neq 0\) \textbf{or} \(C \geq C_{\text{max}}\)}{
        \textcolor{Red}{\textbf{Robot Action:}} Continue or pivot to Red \;
    }
    \Else{
        \textcolor{Red}{\textbf{Robot Action:}} Continue or pivot to Blue \;
    }
    \ForEach{\((x_i, y_i) \in \{(x_i, y_i)\}_{i=1}^N\)}{
        \uIf{\(x_{1b} \leq x_i \leq x_{2b} \land y_{1b} \leq y_i \leq y_{2b}\) \textbf{or} \(x_{1r} \leq x_i \leq x_{2r} \land y_{1r} \leq y_i \leq y_{2r}\)}{
            \textcolor{Red}{\textbf{Robot Action:}} Stop Movement \tcp*{If Human presses any buzzer, then stop and commit to the closest buzzer.}
            \textbf{Break: Exit} 
        }
    }
    
}
$\mathbf{p}_{\text{robot}}$: Current Robot Pose, $d_{\text{blue/red}} = \|\mathbf{p}_{\text{robot}} - \mathbf{p}_{\text{buzzer}}[\text{"blue/red"}][0]\|$\;
\uIf{$d_{\text{blue}} < d_{\text{red}}$}{
    \textcolor{Red}{\textbf{Robot Action:}} Press Blue buzzer \tcp*{Nearest buzzer is blue}
}
\Else{
    \textcolor{Red}{\textbf{Robot Action:}} Press Red buzzer \tcp*{Nearest buzzer is red}
}
\end{algorithm}

\vspace{0.5em}
\noindent \textit{\textbf{Remark 8:}} The Bias Consensus algorithm will terminate when the human and robot opinions reach consensus, i.e., when \(z_r \cdot \hat{z}_h \!>\! 0\). Until this condition is met, the algorithm continuously increases the robot bias \(b_r\) beyond the critical threshold \(b_r^*\), ensuring that the signs of \(z_r\) and \(\hat{z}_h\) become identical at each point during the interaction. Once consensus is achieved, the dynamics ensure that the robot bias \(b_r\) remains in the desired state by continuously monitoring the bias and adjusting the robot's opinion accordingly.
Concurrently, visual cues from the robotic eye are activated to impart bias in human opinion toward the same consensus option \( O \). A detailed mechanism of the gradual increase in visual cues from robotic eyes is discussed in Section \ref{Sec_5}.

Algorithm \ref{algorithm_2} receives \( z_r \) and the set of detected hand landmarks \(\{(x_i, y_i)\}_{i=1}^N\) from Algorithm \ref{algorithm_1}. It then initializes the opinion change vector \( \vec{T} \) and the change count \( C \). To ensure robustness and safety, the robot only alters its actions if every element within \( \vec{T} \) transitions to true and \( C \) exceeds the maximum change threshold \( C_{\text{max}} \). Based on the initial value of \(\text{sgn}(z_r)\), the robot decides its movement action, either towards the red or blue buzzer.

If \(\text{sgn}(z_r)\) transitions from \(1\) to \(-1\) or vice versa, the robot requires \(C_{\text{max}}\) consecutive identical \(\text{sgn}(z_r)\) values, either \(1\) or \(-1\), to confirm a stable directional change before adjusting its trajectory. This mechanism ensures that the robot does not respond to abrupt, transient shifts in human opinion, thereby stabilizing interaction. Once \(C \!\geq\! C_{\text{max}}\), both \( \vec{T} \) and \( C \) are reset to their initial states for the subsequent change evaluation. The robot then pauses for $0.1$ seconds and adjusts its course towards the red or blue option based on the trial number. In trials 1\textsuperscript{st} to 3\textsuperscript{rd}, designed for disagreement, the robot selects the opposite option; in trials 4\textsuperscript{th} to 8\textsuperscript{th}, dynamic bias adjustments override initial disagreements, directing the robot towards the agreed convergence choice \( O \). Upon detection of a hand coordinate within the bounding box of the red or blue buzzer, Algorithm \ref{algorithm_1} ceases the transmission of \( z_r \) and \(\{(x_i, y_i)\}_{i=1}^N\), causing the robot to stop. The robot then commits to press the nearest buzzer, determined by its current end effector position.

\subsection{Selection of Opinion Dynamics Parameters}\label{Sec_4.6}

In this paper, we assume that the key parameters governing the opinion dynamics of the human and robot are identical, allowing both agents to evolve symmetrically within the same decision space and, the system is modeled as a behaviorally homogeneous two-agent, two-option network.

For the numerical analysis in Section \ref{Sec_4.4.1} (a homogeneous two-agent, two-option system with the human and robot as controllable agents), we used the following parameter set for \eqref{eq:1} and \eqref{eq:2}, modeling the human and robot's opinions as agents: \( d_{r} \!=\! d_{h} \!=\! d \!=\! 10, \; u_{r} \!=\! u_{h} \!=\! u \!=\! 2.24, \; \alpha_{r} \!=\! \alpha_{h} \!=\! \alpha \!=\! 0.05, \; \gamma_{r} \!=\! \gamma_{h} \!=\! \gamma \!=\! -8.0, \; a_{rh} \!=\! a_{hr} \!=\! 1 \). Moreover, during experiments, in dissensus trials (1\textsuperscript{st}--3\textsuperscript{rd}) discussed in Section \ref{Sec_4.3} involving dynamic robot opinion, we chose \(\bigl(d_{r}, u_{r}, \alpha_{r}, \gamma_{r}\bigr)\) to match these theoretical values, but with a bias \( b_{r} \!=\! 0 \). In contrast, for the experimental (eye-gaze and biased-opinion) trials (4\textsuperscript{th}--8\textsuperscript{th}) discussed in Section \ref{Sec_4.4.2}, the same robot parameters were retained, but with a \emph{nonzero} bias dynamically updated by \eqref{eq:6}, using parameters \(\lvert \sigma \rvert \!=\! 0.1\) and \(\beta \!=\! -\!1.6\). Finally, for the human-opinion observation discussed Section \ref{Sec_4.5} in during experiments, modeled by \eqref{eq:7}, we set \( a \!=\! 8\) and \(k \!=\! 1.5 \) across all eight trials.

The model parameters listed above were optimized for the experiments through an iterative hit-and-trial approach. These parameters are first used for the theoretical analysis of a two-agent, two-option opinion system and later applied in the human-robot interaction experiment. It is important to note that for any specific human-robot interaction scenario, an optimized set of parameters is required, obtained either through analytical methods or iterative experimentation.

\section{Human Opinion Regulation via Non-Verbal Communication from Robotic Eye-Gaze}\label{Sec_5}
\begin{figure}[ht]
\centering
\includegraphics[width=\linewidth]{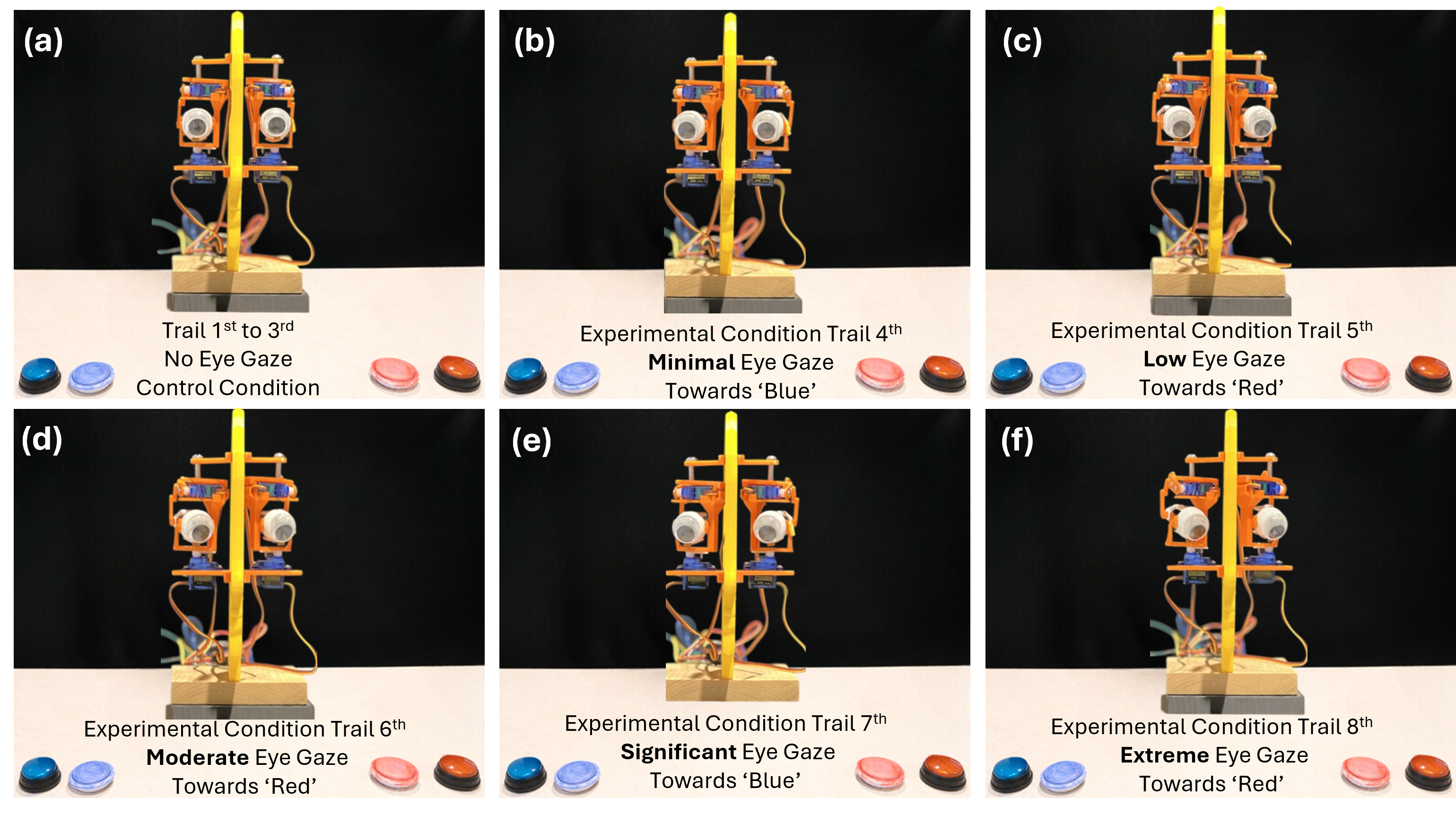}\vspace{-1em}
\caption{Illustration of the human participant's view of the robot eye gaze, robotic arm, and buzzer across different trials. (a) In 1\textsuperscript{st} to 3\textsuperscript{rd} trials, the robot remains neutral, consistently selecting the opposite buzzer color to demonstrate disagreement. (b) In 4\textsuperscript{th} trial, the robot eye gaze is directed towards the blue buzzer, signaling a shift towards alignment. (c) In 5\textsuperscript{th} trial, the robot intensifies its gaze towards the red buzzer (d) In 6\textsuperscript{th} trial, the gaze remains fixed on the red buzzer (e) In 7\textsuperscript{th} trial, the gaze returns to the blue buzzer (f) In 8\textsuperscript{th} trial, the robotic eye gaze intensifies towards the red buzzer, marking the final and most pronounced visual clues. From 4\textsuperscript{th} to 8\textsuperscript{th} trials, the robot progressively strengthens its gaze towards the selected option, affecting human's decision-making process.}
\label{fig:5}
\end{figure}

During the experiment, robotic eyes performed specific movements across different iterations, providing gaze cues to direct participants’ attention toward either the red or blue option. In the first three iterations, the eyes remained centered, looking straight ahead at zero radians to establish a neutral baseline, as shown in Fig.~\ref{fig:5}(a).

In the 4\textsuperscript{th} iteration, the robotic eyes performed a dramatic and attention-grabbing movement by positioning their servos near the mechanical extremes toward the red buzzer, maximizing visual and auditory impact. Following this initial gesture, the eyes then provided a \textbf{minimal} gaze shift at \(-0.47\) radians in yaw (\(\psi\)) and \(0.31\) radians in pitch (\(\theta\)) toward the blue buzzer, as shown in Fig.~\ref{fig:5}(b). In the 5\textsuperscript{th} iteration, the robotic eyes were set to a \textbf{low} gaze setting to enhance visibility, reaching \(+0.53\) radians in yaw toward red and \(-0.53\) radians in pitch downward. Starting from this trial, the common convergence option \(O\) switched from ``blue'' to ``red'' to counteract the expectancy bias from the 4\textsuperscript{th} trial. By altering the gaze direction, the experiment tested whether participants would recalibrate their expectations and decisions when confronted with inconsistent cues, thus mitigating the influence of prior experiences on their current perceptions and choices. During the 6\textsuperscript{th} iteration, the robotic eyes were set to a \textbf{moderate} gaze setting, reaching \(+0.62\) radians in yaw and \(-0.62\) radians in pitch downward. This adjustment increased the visual prominence of the gaze, guiding participants to rely more on immediate visual cues for decision-making rather than pattern-based expectations.

In the 7\textsuperscript{th} iteration, the robotic eyes were set to a \textbf{significant} gaze, shifting to \(-0.72\) radians in yaw and \(+0.72\) radians in pitch, redirecting attention toward the blue buzzer. In the 8\textsuperscript{th} iteration, the gaze reached an \textbf{extreme} setting, positioning the eyes at approximately \(+1.09\) radians in yaw and \(-0.94\) radians in pitch, creating the most pronounced gaze towards the red buzzer.

Additionally, an experimental study \cite{Nakao2013} found that even in the human brain’s resting state when not engaged with direct stimuli, it can still prepare for and influence subsequent decision-making. This foundational activity was evident in trials 1\textsuperscript{st} to 3\textsuperscript{rd} of our study, where despite the robotic eye gaze being neutral and offering no external cues, participants' decisions were subtly shaped by their intrinsic brain activity. These internal biases may have served as initial conditions for dynamic human opinion formation. In contrast, during trials 4\textsuperscript{th} to 8\textsuperscript{th}, the activation of the robotic eye gaze just before the onset of auditory cues, ``1, 2, 3, go", marked a shift to externally guided decision-making. Here, participants' brain activity may have been more dynamically responsive to these immediate external stimuli rather than being predominantly influenced by the brain's resting state, effectively mitigating internal bias. The presence of directed visual cues from the robotic eye shifted the cognitive processing from an internally guided to an externally responsive mode, aligning with findings that external stimuli can override internal predispositions in decision contexts.

Thus, in early trials, internal biases define the initial conditions for human opinion dynamics, which are updated by observing the robot's behavior, as evidenced by changes in intended movements of participants. In later trials, the robotic eye provides an external bias by indicating which option the human should choose, thus mitigating internal biases. Consequently, participants respond more directly to observed cues, aligning with the non-linear opinion dynamics in \eqref{eq:2}. Furthermore, since participants were not explicitly informed about the purpose and functioning of robotic eyes at any point before or during the experiment and had to deduce environmental cues on their own, this approach effectively prevented any preconceived or expected bias.

\section{Experimental Demonstrations}\label{Sec_6}
\begin{figure}[p]
\centering
\includegraphics[width=\linewidth]{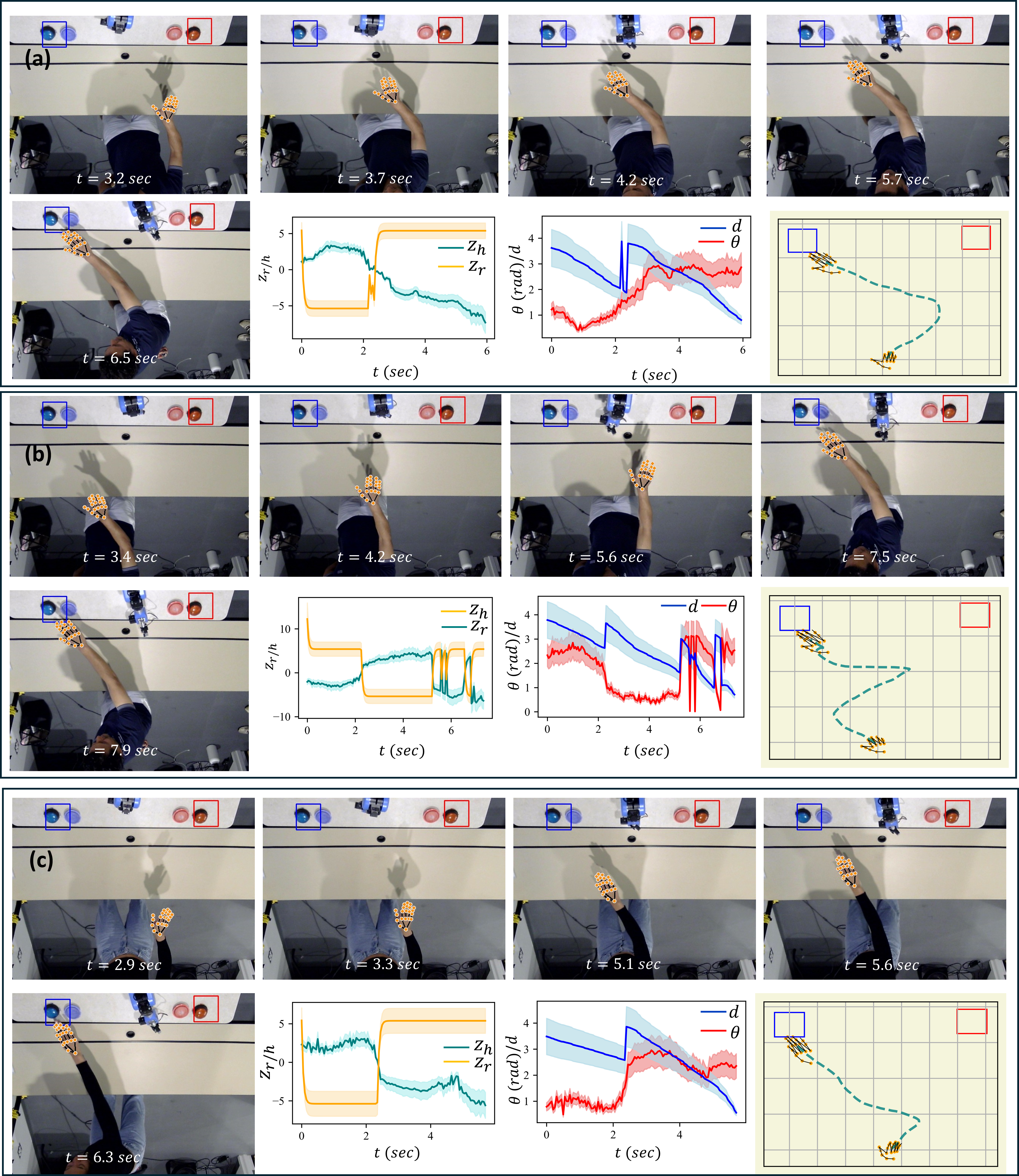}\vspace{-0.5em}
\caption{Experimental illustration of mid-switch, multiple switches, and strategic early switch by human participants in trials 1\textsuperscript{st} to 3\textsuperscript{rd}, where the robot is configured to show disagreement.}
\label{fig:6}
\end{figure}

For the presented human-robot decision-making experiment, Algorithms \ref{algorithm_1} and \ref{algorithm_2} were extensively evaluated to ensure robustness and reproducibility in robot behavior, particularly in handling individual variability in participants’ actions and beliefs, as well as generalizability across the entire cohort of $51$ human participants. Fig.~\ref{fig:6} illustrates the participant interactions during the first three disagreement trials. In these trials, Algorithm \ref{algorithm_2} ensured that the robot consistently selected a buzzer color contrary to the participant’s initial choice. Fig.~\ref{fig:6}(a) illustrates the interaction dynamics during the first trial of the $32^{\text{nd}}$ participant. Initially inclined to press the red buzzer, the participant updated their opinion $z_h$ to align with the robot's intended choice after observing it moving toward the blue buzzer at \(t \!=\! 3.7\) seconds. Simultaneously, operating under Algorithm \ref{algorithm_1}, the robot proactively altered its trajectory following the directives of Algorithm \ref{algorithm_2}. The distance \(d\) decreased progressively as the human neared a decision, and the angle \(\theta\) was adjusted accordingly to reflect the human's updated opinion. An accompanying plot of the human hand trajectory highlights the mid-interaction modification of human opinion.

At the beginning of the experiment, all participants were instructed that they were allowed a single alteration in their path before crossing the black decision commit line to match the robot's choice. Any violation of this rule would result in a one-point penalty deducted from the participant’s overall score. Despite these guidelines, some participants changed their paths multiple times and attempted to align their choices with those of the robot, thereby attempting to maximize their scores. Fig.~\ref{fig:6}(b) illustrates $32^{\text{nd}}$ participant in $3^{\text{rd}}$ trial after experiencing repeated disagreements from the robot in first two trials. Initially heading towards the blue buzzer, the participant switched to red, then reverted to blue, and made a last-minute attempt to switch back to red upon observing the robot's movement. However, upon the realization that he had already crossed the decision commit line, he ultimately reverted and committed to blue. This scenario demonstrates the robot’s deceptive behavior capability, driven by Algorithm \ref{algorithm_2}, alongside its ability to rapidly form reliable disagreeing opinions using non-linear opinion dynamics. The robot adapts to multiple directional changes in response to varying human opinions and actively counters human actions with multiple switches.

Fig.~\ref{fig:6}(c) illustrates the $47^{{th}}$ participant in the $3^{rd} $ trial, with the robot configured for disagreement. Having previously faced two trials of deliberate opposition from the robot, the participant anticipated further adversarial robot behavior. In response, the participant initiated an early strategic switch at the start of the interaction. However, the robot's swift and adaptive opinion modifications, enabled by Algorithm \ref{algorithm_1}, allowed it to quickly adjust its responses to continue its disagreement. This scenario effectively simulates real-world conditions in which continuous disagreement necessitates adaptive collaboration methods. These trials underscore the need for advanced collaborative mechanisms that lead to the deployment of robotic eye gaze and bias control algorithms as facilitators to achieve consensus between human and robot decisions.

\begin{figure}[ht]
\centering
\includegraphics[width=\linewidth]{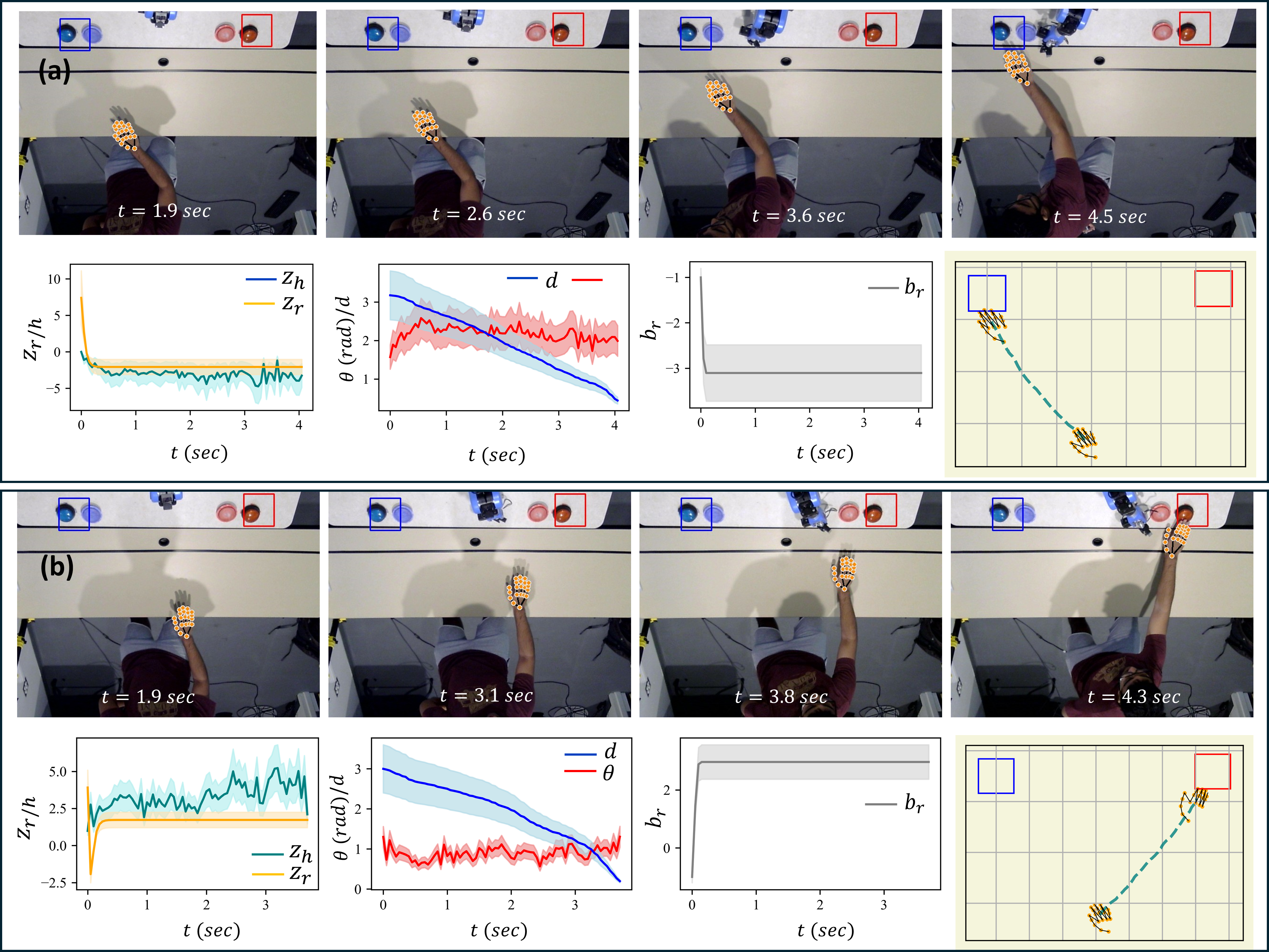}\vspace{-0.5em}
\caption{Illustration of trials 9\textsuperscript{th} and 10\textsuperscript{th}, where the robotic eye is activated to provide visual cues to the human participant, showing no adjustment in hand path trajectory and a direct consensus between human and robot actions as well as opinions.}
\label{fig:7}
\end{figure} 

Fig.~\ref{fig:7}(a) displays $35^{\text{th}}$ participant in the $7^{\text{th}}$ trial, where the robotic eye was activated prior to the interaction, directing its gaze toward the blue buzzer, as shown in Fig.~\ref{fig:5}(e). The bias control algorithm within Algorithm \ref{algorithm_1} was activated, with the non-linear robot opinion, bias dynamics, and human opinion observation parameters set according to Table 1. With input \(O \!=\! ``blue"\), the robot's opinion was directed toward the blue buzzer, increasing its bias toward this selection. The bias plot in Fig.~\ref{fig:7}(a) indicates that once the robot’s bias (\(b_r\)) exceeded the critical threshold, which indicates social attention to human opinion, it consistently maintained this bias toward the blue option for the rest of the interaction. Meanwhile, the human’s internal opinion dynamics were influenced by the bias imparted from the robot’s eye gaze, compelling the participant to firmly press the blue buzzer without any deviations in hand trajectory. In the 8\textsuperscript{th} and final trial of the same $35^{\text{th}}$ participant, the robotic eye was oriented toward the red buzzer, with Algorithm \ref{algorithm_2} configured for the input \(O \!=\! ``red"\). This final interaction, shown in Fig.~\ref{fig:7}(b), demonstrated the most pronounced effects of eye gaze and recorded the shortest time to consensus among all trials. This outcome illustrates that as participants internalize the external stimulus or bias from the robotic eye—and as the robot consistently aligns with the same buzzer color—their trust in the robotic guidance intensifies, leading to a quicker consensus.

Although most trials configured for disagreement resulted in dissensus, there were instances where quick, strategic end-moment switches enabled some participants to align their choices with the robot's final choices during the initial trial. However, almost all participants encountered at least one instance of disagreement, which motivated them to actively observe the environment for guidance cues from the robotic eye gaze, thereby facilitating collaboration. Furthermore, in the early rounds, when the robotic eye was activated, participants—driven by growing trust in the robot's behavior—adjusted their hand paths to align with the robot's choices. The progression of these switches across the interaction trials is thoroughly discussed in Section \ref{Sec_7}, which presents the comprehensive results of all experiments and the overall trends.

\section{Experimental Outcomes}\label{Sec_7}
We categorized each participant's decisions into four nominal outcomes: \textbf{Consensus (C)} if no path alteration occurred and the choice matched the robot's, \textbf{Dissensus (D)} if no path alteration occurred but the choice differed from the robot's, \textbf{Consensus with Hesitation (CH)} if exactly one hand-path modification was made toward agreement, and \textbf{Dissensus with Hesitation (DH)} if a single path change led to final disagreement. Participants then progressed through two main phases: (1) a control condition (Trials~1\textsuperscript{st}---3\textsuperscript{rd}), during which the robot provided no non-verbal communication through eye gaze and maintained an unbiased opinion, and (2) an experimental condition (Trials~4\textsuperscript{th}---8\textsuperscript{th}) involving non-verbal communication through robotic eye gaze and a biased robot opinion favoring a particular choice.
\begin{figure}[t]
\centering
\includegraphics[width=\linewidth]{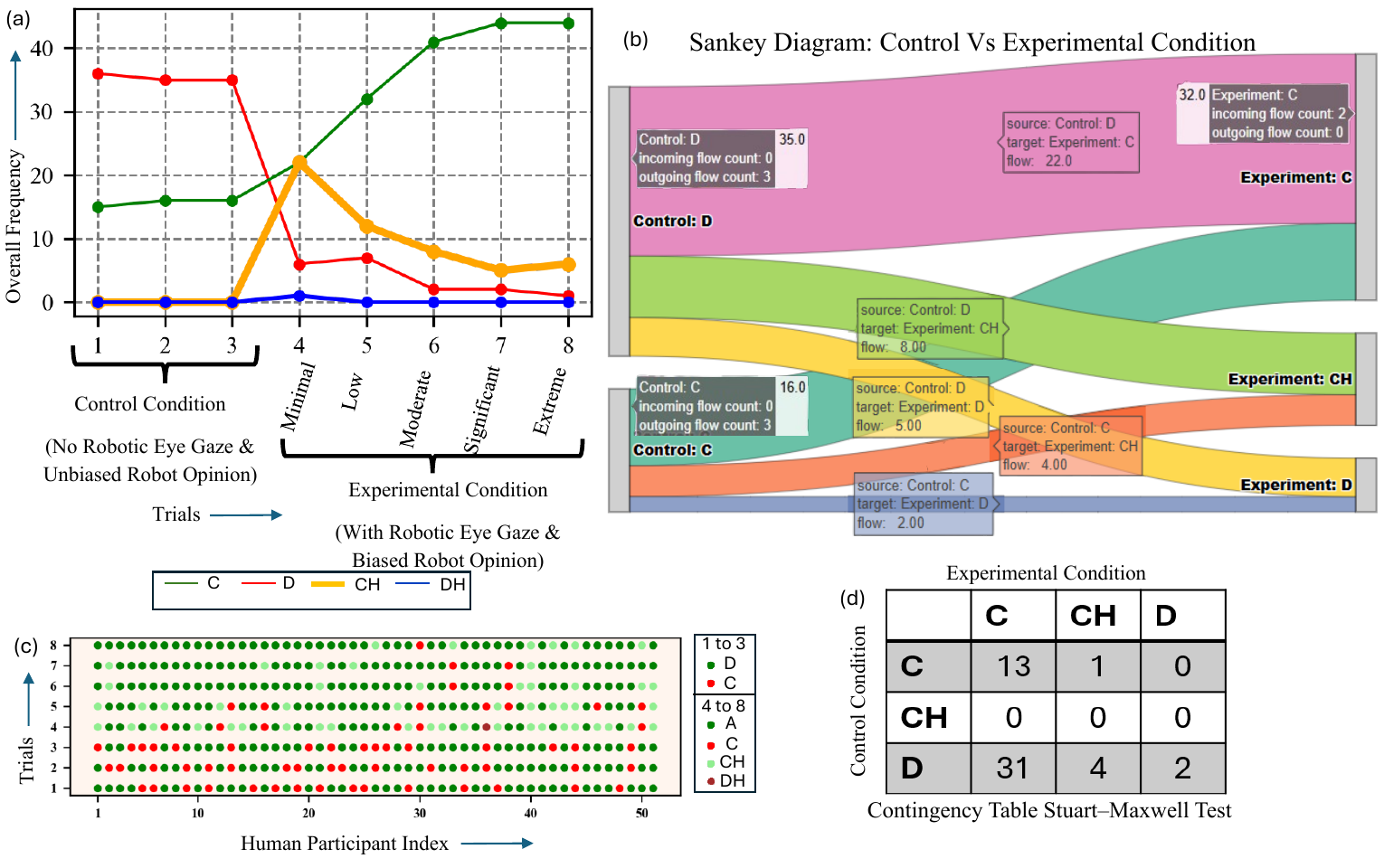}\vspace{-1em}
\caption{(a) Frequency distribution across trials, showing transition from dissensus to consensus when robotic gaze and biased opinions were introduced. (b) Sankey diagram depicting participant flow between decision categories from control to experimental conditions. (c) Individual participant responses across all trials. (d) Contingency table with Stuart-Maxwell test results confirming significant shift toward consensus.}
\label{fig:8}
\end{figure} 
     
\subsection{Consensus}\label{sec_7.1}
The distribution of participants’ nominal outcomes (\(C\), \(CH\), \(D\), and \(DH\)) across eight trials is depicted in Fig.~\ref{fig:8}(a). In the first three trials, where the robot adopted a dissensus stance, \(D\) (dissensus) prevailed at around 70\% of choices on average (\(70.59\%\) in Trial~1\textsuperscript{st} and \(68.63\%\) in Trials~2\textsuperscript{nd}--3\textsuperscript{rd}). By contrast, \(C\) (consensus) was near \(30\%\), and nearly no \(CH\) (consensus-with-hesitation). No participant selected \(DH\) (dissensus-with-hesitation) in these early trials, suggesting a strong preference against disadvantageous outcomes. Beginning in Trial~4\textsuperscript{th}, once the robot’s eye-gaze cues became salient, \(D\) (dissensus) dropped from about \(12\%\), \(14\%\) in Trials~4\textsuperscript{th}--5\textsuperscript{th}, down to below \(4\%\) in Trials~6\textsuperscript{th}--7\textsuperscript{th}, and ultimately reaching \(1.96\%\) by Trial~8\textsuperscript{th}. Simultaneously, \(C\)  (consensus) increased from \(43.14\%\) in Trial~4\textsuperscript{th} to nearly \(86\%\) by Trials~7\textsuperscript{th}--8\textsuperscript{th}. Meanwhile, \(CH\) (consensus-with-hesitation) emerged at about \(20\%\) on average through Trials~4\textsuperscript{th}--8\textsuperscript{th}, reaching \(11.76\%\) in the final trial. Overall, \(C + CH\) exceeded \(90\%\) by Trial~8\textsuperscript{th}, highlighting a pronounced shift toward consensus-based choices. In sum, these robot-led cues nudged participants toward collaboration rather than dissensus: Trials~1\textsuperscript{st}--3\textsuperscript{rd} featured \(69\%\) dissensus versus \(1\%\) consensus, whereas Trials~4\textsuperscript{th}--8\textsuperscript{th} reversed that pattern to \(7\%\) dissensus versus \(92\%\) consensus-oriented responses.

Fig.~\ref{fig:8}(b) shows a Sankey diagram showing how participants moved from each \emph{Control} category (on the left) to each \emph{Experiment} category (on the right). The width of each colored flow depicts the number of participants transitioning between categories. For instance, among the $35$ participants who were classified as \(D\) (dissensus) during the Control phase, $22$ ultimately shifted to \(C\) (consensus) under the Experiment condition, $8$ switched to \(CH\) (consensus-with-hesitation), and $5$ remained in \(D\). Similarly, out of the $16$ participants who began in \(C\) during the Control phase, $4$ moved to \(CH\), $2$ moved to \(D\), and so forth. By examining the node labels and flow widths, one can pinpoint exactly how many participants changed or maintained their categories. Notably, the \emph{largest flow} occurs from Control \(D\) to Experiment\(D\), indicating that a sizable subset of participants initially aligned with dissensus were drawn to consensus once the experimental manipulation (robotic eye gaze and biased robot opinion) took effect. Overall, the Experiment phase exhibits a stronger clustering of participants in \(C\), and to a lesser degree \(CH\), illustrating a shift toward more unified or “consensus-like” outcomes rather than persisting in the more widely distributed Control categories.

To statistically confirm that participants’ final choices differed between the Control and Experiment conditions, we performed a \textbf{Stuart--Maxwell test} for marginal homogeneity. The statistical test using the contingency table shown in Fig.~\ref{fig:8}(d) yielded a test statistic of $68.125$ with a corresponding $p$-value below $10^{-8}$. In other words, there is an \emph{exceptionally strong difference} in the marginal distributions across the two phases. Consequently, these results provide \textbf{robust evidence} that the observed rise in \emph{consensus} (\(C\) or \(CH\)) is significantly influenced by the introduction of \emph{non-verbal robotic eye gaze} and a \emph{deliberately biased robot opinion} in the collaboration task. These findings provide strong support for the acceptance of hypothesis \textbf{H1} and address \textbf{RQ1}, confirming that non-verbal communication through robotic eye-gaze, combined with a biased robot opinion, facilitates consensus in human-robot co-learning.

\subsection{Trust}\label{sec_7.2}
At \emph{Trial~4\textsuperscript{th}} (\textit{Minimal} eye-gaze intensity), a pronounced spike in \(CH\) (consensus-with-hesitation) arose, coinciding with the robot’s abrupt shift from dissensus to bias assimilation. This sudden behavioral change prompted participants to scrutinize the newly activated robotic eye. The audible servo noises and mysterious purpose of its gaze led them to adopt a cautious ``try-and-see'' approach: \(CH\) and \(C\) (full consensus) each accounted for roughly $43\%$ of choices. Over subsequent trials, as the eye-gaze setting increased from \textit{Low} (Trial~5) to \textit{Moderate} (Trial~6) and eventually \textit{Significant} (Trial~7) and \textit{Extreme} (Trial~8), \(CH\) diminished to $11.76\%$ while \(C\) climbed to $86.27\%$. By Trial~8, \(D\) (dissensus) stood at only $1.96\%$, suggesting that growing trust in the robot’s cues supplanted both hesitation and dissensus.

We again performed a Stuart--Maxwell test on each pair of adjacent trials, which are \emph{Trial~4\textsuperscript{th}} to \emph{Trial~5\textsuperscript{th}},  \emph{Trial~5\textsuperscript{th}} to \emph{Trial~6\textsuperscript{th}}, \emph{Trial~6\textsuperscript{th}} to \emph{Trial~7\textsuperscript{th}}, and \emph{Trial~7\textsuperscript{th}} to \emph{Trial~8\textsuperscript{th}}, to see how the nominal outcomes \((C,CH,D,DH)\) evolved over time. 
Between \emph{Trial~4} and \emph{Trial~5}, the test yielded \(\chi^2 \!=\! 12.39\), \(p \!=\! 0.0062\), showing a significant shift---predominantly from \(CH\) (consensus-with-hesitation) and \(D\) (dissensus) toward \(C\) (consensus). Likewise, \emph{Trial~5} to \emph{Trial~6} also showed a significant change (\(\chi^2 \!=\! 9.48\), \(p \!=\! 0.0087\)), indicating participants were rapidly discarding hesitation as they verified the robot's cues. 
However, by \emph{Trial~6} to \emph{Trial~7} (\(\chi^2 \!=\! 2.57\), \(p \!=\! 0.277\)) and \emph{Trial~7} to \emph{Trial~8} (\(\chi^2 \!=\! 0.44\), \(p \!=\! 0.801\)), no statistically significant redistribution of outcomes occurred, suggesting that most participants had already transitioned into a stable, trust-driven consensus. Overall, these statistical results provide strong evidence that \emph{once the robot's guidance was perceived as reliable}, participants' \emph{trust} soared, driving a rapid convergence on full consensus \((C)\). Rather than remaining in a state of hesitation \((CH)\) or dissensus \((D)\), the majority of participants ultimately aligned confidently with the robot's cues, indicating that effective robotic feedback can overcome initial skepticism and promote stable, trust-based collaboration. These progressive changes in participant behavior across trials provide strong support for the acceptance of hypothesis \textbf{H2}. Addressing \textbf{RQ2}, the results confirm that participants’ trust in robotic guidance can progressively develop through increased  exposure to consistent non-verbal cues.

Furthermore Fig.~\ref{fig:8}(c) presents the comprehensive data from the experiment with $51$ participants across all trials. Hypothetically, assuming perfect disagreement in the initial three trials---with participants strictly adhering to the experimental protocol without making last-minute changes or quick switches---and assuming hypothetical agents (human participants) fully trusted the robotic eye cues from the $4^{\text{th}}$ to $8^{\text{th}}$ trials, the scatter plot should predominantly show all green points, indicating disagreement initially and agreement subsequently. Contrary to expectations, the unpredictability of irrational human behavior led to significant deviations. Red points highlight unexpected agreements in the initial trials and disagreements in the later ones. Each participant only exhibited a change to agreement behavior starting from the $4^{\text{th}}$ trial, which gradually diminished, transitioning to complete agreement by the $8^{\text{th}}$ iteration. The majority of the $51$ participants reached a consensus in the last five trials, with only a few outliers. The potential reasons for these outliers, including participants' educational backgrounds, are discussed in Section~\ref{Sec_8}.

\section{User Feedback Analysis}\label{Sec_8}
In addition to collecting quantitative data during the experiment, we also gathered post-experiment feedback and conducted brief interviews to capture participants’ subjective impressions. Each participant answered the following five questions, either using a $10$-point Likert scale or choosing between “Yes,” “No,” or “Maybe,” as appropriate:

\begin{enumerate}
    \item \emph{Did you feel the robot was working with you or against you during the first three (initial) trials?} 
    (1 = Working against me, 10 = Working with me)
    
    \item \emph{Did you take visual hints provided by the robotic eye?} 
    (1 = Not at all, 10 = Very heavily)
    
    \item \emph{Did the clicker placed behind you cause a split in your attention or slow your responses?} 
    (Yes/No/Maybe)
    
    \item \emph{Did you perceive that the robotic eye’s hints varied noticeably across iterations, becoming more pronounced in the final trials?}
    (1 = Not at all, 10 = Very heavily)
    
    \item \emph{By the end (trials 4–8), did you feel the robot was working collaboratively with you?}
    (1 = Working against me, 10 = Working with me)
\end{enumerate}

We conducted one-sample \emph{t}-tests on each numeric question (Q1, Q2, Q4, Q5), comparing their means against the neutral midpoint of $5.5$ on a $10$-point Likert scale. We also performed a chi-square goodness-of-fit test for Q3 (a categorical Yes/No/Maybe question). For Q1, with a mean of $3.49$ and a test statistic of $t \!=\! -6.831$ ($p \!<\! 10^{-4}$), scores were significantly below the neutral midpoint. Hence, participants perceived that, in the initial trials, the robot was \emph{not} working in tandem with them but instead seemed to act “against” their interests. This designed ``dissensus'' behavior was intended to elicit greater collaborative effort and attentiveness from human participants. The results further confirm that non-linear opinion dynamics can facilitate fast, flexible decision-making, enabling the robot to adopt a controlled dissensus state whenever needed.

The mean rating for Q2 was $6.86$, significantly above $5.5$ ($t \!=\! 2.565$, $p \!=\! 0.0134$), indicating that participants did, overall, notice and utilize the robotic eye’s hints to a moderate or substantial degree. Regarding the shift in eye-gaze prominence (Q4) and perceived collaboration (Q5), Q4 had a mean of $6.41$ ($t \!= \!2.251$, $p \!=\! 0.0288$), suggesting participants found the robot’s eye-gaze cues to become more pronounced in later trials. Meanwhile, Q5 yielded the highest mean (8.35) with $t \!=\! 12.370$ ($p \!<\! 10^{-4}$), reflecting a strong consensus that, by the final phase, the robot was decidedly “working with” participants. Taken together, these findings point to a clear escalation in cooperative cues and an ultimately high sense of collaboration—consistent with prior evidence that eye-gaze can effectively foster consensus. In investigating the cognitive load and attentional split (Q3), the responses were not evenly distributed among “Yes,” “No,” and “Maybe,” with $34$ “Yes,” $9$ “Maybe,” and $8$ “No” ($\chi^2 \!=\! 11.368$, $p \!=\! 0.0034$). A significant majority of participants therefore reported that placing the clicker behind them \emph{did} cause an attentional split or slowed their responses. In general, these statistical results demonstrate that, although participants initially felt the robot was working against them, they eventually accepted and used its hints, particularly as the gaze cues became more prominent. In the end, the participants overwhelmingly regarded the robot as supportive, aligning with the high Q5 scores and the observed shift towards consensus-based behaviors.

Additionally, from the post-experiment verbal interviews, it was observed that participants from robotics backgrounds predominantly perceived the robotic eyes as camera sensors intended to detect their actions, even though the robotic eyes lacked any camera functionality and were designed to guide participants' decisions toward consensus with the robot. In contrast, participants with general engineering backgrounds often interpreted robotic eyes as mechanisms intended to confuse them and negatively affect their scores, a perception potentially influenced by the initial three trials configured for disagreement. In contrast, participants from non-technical backgrounds, such as maintenance workers, typically viewed the robotic eyes as a straightforward visual guidance source, enabling effective collaboration with the robot without forming any misconceptions. In addition, some participants did not trust the robotic eye’s hints due to the robot’s initial dissensus behavior, suspecting it was intentionally trying to mislead them. A small subgroup explicitly revealed in interviews that, following the early disagreement phase, they continued perceiving the non-verbal cues as potentially deceptive. Yet, in our study—where the worst-case scenario had the robot begin in dissensus—most participants ultimately shifted from viewing the robot as ``tricky” to regarding it as a genuinely \emph{collaborative} partner once they discovered the eye-gaze signals were reliable. Consequently, one can infer that if initial interactions had instead been cooperative \emph{without} eye gaze, then introducing eye-gaze cues in later stages would likely foster even stronger trust and more seamless collaboration. Establishing trust from the outset thus appears crucial for ensuring that non-verbal robot communication is interpreted positively rather than suspiciously.

\section{Conclusion}\label{Sec_9}
This work demonstrates how effective consensus and trust can develop between a human and a robot by employing nonverbal communication---specifically, robotic eye gaze as an external bias or stimulus, governed by nonlinear opinion dynamics. Our experimental findings and participant feedback both confirm that visually based cues (e.g., an increasingly pronounced robotic eye gaze) can guide interactions from dissensus to consensus. In particular, as the robot’s eye gaze becomes clearer, participants better perceive the robot’s intent and more reliably align with the robot’s choices, thereby increasing both consensus rates and trust over successive trials.

We also introduced a nonlinear opinion-dynamics model with a novel dynamic bias framework, supported by numerical parameter sweeping and equilibrium analysis. In this proposed model, both human and robot biases must point to the same option in a two-choice setting and exceed the agents’ inherent social attentions to guarantee convergence toward a unique consensus equilibrium. This theoretical insight underpins our experimental design, illustrating how appropriately tuned external stimuli (i.e., robotic eye cues) can overcome initial disagreement and ultimately yield a stable, trust-driven consensus.


\section{Limitations and Future Work}\label{Sec_10}

Although our experimental study and bias-controlled opinion-dynamics model successfully demonstrate two-choice consensus-building, many real-world contexts involve multiple options. For example, a human--robot search-and-rescue team might need to decide which of several buildings to prioritize, or a scheduling robot could manage multiple manufacturing tasks with different deadlines. Adapting the current model to handle multi-option scenarios requires extending how biases and social attentions converge on a sequence of choices, rather than just a binary outcome. 

Furthermore, even though our participants came from diverse backgrounds, the overall setup was still a controlled laboratory environment with a single robot arm, a static workspace, and a specific scoring incentive. Deploying and validating this approach in uncontrolled industrial or public settings---characterized by noise, added safety constraints, and strict time demands---remains an essential step. Real-world contexts may also necessitate adaptive bias weighting that responds to the user's attention, fatigue, or evolving expertise. Our experiments only tuned the robot's bias parameters for one scenario; broader generalization or automated parameter identification must be explored to accommodate varied tasks and user profiles.

Additionally, our model only partially addresses the potential divergence between implicit (internal) and explicit (outwardly displayed) opinions \cite{Zhang2024}, which can create discrepancies when external social pressures influence individuals' actions. We have also not formally incorporated trust into opinion dynamics \cite{Jiang2024}, although trust can critically shape how biases alter agent decisions. Future work will thus extend the bias-control framework to multi-choice situations, differentiate implicit from explicit opinions, and quantify trust---all vital for a more comprehensive, real-world--ready model of human--robot collaboration. In parallel, our goal is to develop more realistic models of bias and opinion formation that capture evolving human behaviors under uncertain, real-world conditions. We also plan to create a control algorithm for robot behavior that accounts for human cognitive complexities and dynamic trust. Future experimentation will move toward uncontrolled environments such as warehouses and factories, where workers interact daily with robots under high cognitive load. We also envision adaptive bias-tuning methods that automatically adjust both the robot's and the user's parameters (through robotic cues) based on real-time performance, perceived trust, and workload. Finally, we will investigate multi-modal communication channels beyond gaze (e.g., haptic or auditory signals, AR interfaces) to see if they further enhance consensus and diminish hesitation.


\end{document}